\documentclass[pdflatex,sn-mathphys-num]{sn-jnl}

\usepackage{graphicx}%
\usepackage{multirow}%
\usepackage{amsmath,amssymb,amsfonts}%
\usepackage{amsthm}%
\usepackage{mathrsfs}%
\usepackage[title]{appendix}%
\usepackage{xcolor}%
\usepackage{textcomp}%
\usepackage{manyfoot}%
\usepackage{booktabs}%
\usepackage{algorithm}%
\usepackage{algorithmicx}%
\usepackage{algpseudocode}%
\usepackage{listings}%
\usepackage{adjustbox} 
\usepackage{makecell}

\theoremstyle{thmstyleone}%
\theoremstyle{thmstyletwo}%

\theoremstyle{thmstylethree}%

\begin{document}

\title[Article Title]{TeleEval-OS: Performance evaluations of large language models for operations scheduling}

\author [2]{\fnm{Yanyan} \sur{Wang}} \email{wang.yanyan@ustcinfo.com}

\author *[2]{\fnm{Yingying} \sur{Wang}} \email{wang.yingying@ustcinfo.com}

\author[1]{\fnm{Junli} \sur{liang}} \email{jlliang@mail.ustc.edu.cn}

\author[2]{\fnm{Yin} \sur{Xu}} \email{xu.yin@ustcinfo.com}

\author[1]{\fnm{Yunlong} \sur{Liu}} \email{liuyunlong@mail.ustc.edu.cn}

\author[1]{\fnm{Yiming} \sur{Xu}} \email{xym30@mail.ustc.edu.cn}

\author[1]{\fnm{Zhengwang} \sur{Jiang}} \email{jzw02@mail.ustc.edu.cn}

\author[1]{\fnm{Zhehe} \sur{Li}} \email{lizh2002@mail.ustc.edu.cn}

\author[1,2]{\fnm{Fei} \sur{Li}} \email{fli312@mail.ustc.edu.cn}

\author[1,2]{\fnm{Long} \sur{Zhao}} \email{lzhao@ustcinfo.com}

\author[1,2]{\fnm{Kuang} \sur{Xu}} \email{kxu@ustcinfo.com}

\author *[1]{\fnm{Qi} \sur{Song}}\email{qisong09@ustc.edu.cn}

\author [1]{\fnm{Xiangyang} \sur{Li}} \email{xiangyangli@ustc.edu.cn}

\affil *[1]{\orgdiv{School of Computer Science and Technology}, \orgname{University of Science and Technology of China}, \orgaddress{\city{Hefei}, \postcode{230022}, \state{Anhui}, \country{China}}}

\affil[2]{\orgdiv{Innovation + Research Institute}, \orgname{GuoChuang Cloud Technology Ltd.}, \orgaddress{\city{Hefei}, \postcode{230031}, \state{Anhui}, \country{China}}}


\abstract{The rapid advancement of large language models (LLMs) has significantly propelled progress in artificial intelligence, demonstrating substantial application potential across multiple specialized domains. Telecommunications operation scheduling (OS) is a critical aspect of the telecommunications industry, involving the coordinated management of networks, services, risks, and human resources to optimize production scheduling and ensure unified service control. However, the inherent complexity and domain-specific nature of OS tasks, coupled with the absence of comprehensive evaluation benchmarks, have hindered thorough exploration of LLMs' application potential in this critical field. To address this research gap, we propose the first Telecommunications Operation Scheduling Evaluation Benchmark (TeleEval-OS). Specifically, this benchmark comprises 15 datasets across 13 subtasks, comprehensively simulating four key operational stages: intelligent ticket creation, intelligent ticket handling, intelligent ticket closure, and intelligent evaluation. To systematically assess the performance of LLMs on tasks of varying complexity, we categorize their capabilities in telecommunications operation scheduling into four hierarchical levels, arranged in ascending order of difficulty: basic NLP, knowledge Q\&A, report generation, and report analysis. On TeleEval-OS, we leverage zero-shot and few-shot evaluation methods to comprehensively assess 10 open-source LLMs (e.g., DeepSeek-V3) and 4 closed-source LLMs (e.g., GPT-4o) across diverse scenarios. Experimental results demonstrate that open-source LLMs can outperform closed-source LLMs in specific scenarios, highlighting their significant potential and value in the field of telecommunications operation scheduling.}

\keywords{Benchmark, Large language model, Telecommunications operation scheduling domain}



\maketitle

\section{Introduction}\label{sec1}

In recent years, large language models (LLMs), represented by GPT-4~\cite{achiam2023gpt}, DeepSeek~\cite{liu2024deepseek}, Qwen ~\cite{yang2024qwen2}, and LLaMA~\cite{touvron2023open}, have experienced rapid advancement, demonstrating exceptional capabilities in language comprehension and generation. These models have demonstrated outstanding performance in tasks such as natural language processing, knowledge-based question answering, and content generation. Furthermore, their potential applications have been actively explored in domains such as healthcare~\cite{santorinaios2024medguard, ye2025gmai}, finance~\cite{zhu2024benchmarking, krumdick2024bizbench}, and education~\cite{chen2024dr, liu2024comet}, offering new opportunities for the intelligent transformation of industries. As key providers of information and communication infrastructure, telecommunications operators are also actively exploring the application potential of these large models in real-world business scenarios to enhance service efficiency and intelligence levels~\cite{wang2024automatic}.

As LLMs gradually permeate various facets of telecommunications operations, effectively measuring the performance of LLMs in complex scenarios has emerged as a critical challenge requiring urgent resolution. Currently, numerous evaluation benchmarks have been developed to assess the performance of LLMs across different tasks and domains. These benchmarks can be broadly classified into two categories: general-domain and vertical-domain benchmarks. General-domain benchmarks, such as PromptBench~\cite{zhu2024promptbench}, SG-Bench~\cite{mou2025sg}, and CMMLU~\cite{li2023cmmlu}, are designed to evaluate a broad range of capabilities, including multi-task performance, multilingual proficiency, and domain-specific knowledge. On the other hand, vertical-domain benchmarks focus on specific industries, such as MedExQA~\cite{kim2024medexqa} and TCMBench~\cite{yue2024tcmbench} in healthcare, FinBen~\cite{xie2025finben} in finance, and LAiW~\cite{dai2025laiw} in law. However, due to the inherent complexity and scenario-specific nature of tasks in the telecommunications sector, these existing benchmarks are inadequate for effectively assessing the performance of LLMs in real-world telecommunications scenarios~\cite{bariah2024large, maatouk2024large}.

Telecommunications operational scheduling refers to the coordinated management of networks, services, risks, and personnel resources to optimize production scheduling and achieve unified service control. Its core capabilities include critical processes such as cloud network fault resolution, customer service assurance, major risk operations, and production command scheduling. The unique nature of tasks in this domain poses significant challenges for existing evaluation benchmarks to effectively assess the performance of LLMs in these scenarios. First, as illustrated in Figure 1, the domain features complex scenarios and diverse tasks, which can be categorized into four principal groups: intelligent ticket creation, intelligent handling, intelligent ticket closure, and intelligent evaluation. These groups include a range of tasks, such as homogeneous ticket recommendation, multi-tier intent recognition, and ticket closure report generation. Second, historical tickets contain substantial low-quality data characterized by noisy entries and incomplete records. Finally, ticket data frequently includes sensitive information (e.g., broadband account details, physical addresses) that requires strict data desensitization processes. Therefore, constructing an evaluation benchmark specifically tailored to telecommunications operational scheduling scenarios is of paramount importance~\cite{li2024performance}.

To address this gap, we present TeleEval-OS, a specialized benchmark designed to evaluate LLM capabilities in telecommunications operations dispatching. First, we divide the scenarios into 13 subtasks based on the four stages of operational scheduling business: intelligent ticket creation, intelligent handling, intelligent ticket closure, and intelligent evaluation. Concurrently, we define four capability categories: basic NLP, knowledge Q\&A (Question and Answering), report generation, and report analysis. Subsequently, in accordance with the unique characteristics of operational scheduling tasks, we constructed an evaluation dataset comprising over 10.4k examples. Finally, we evaluated various LLMs' performance in operational scheduling scenarios.

The primary contributions of this research can be summarized as follows:
\begin{itemize}
    \item This paper presents TeleEval-OS, the first comprehensive evaluation framework designed to assess the capabilities of various LLMs across four critical scenarios: intelligent ticket creation, intelligent handling, intelligent ticket closure, and intelligent evaluation. This framework establishes a unified evaluation standard for the intelligent development of telecommunications operational scheduling.
    \item We have constructed an extensive dataset comprising 15 datasets across 13 subtasks, with over 10.4k evaluation samples, covering core capabilities such as basic NLP, knowledge Q\&A, report generation, and report analysis. This provides scientific support for evaluating the performance of LLMs in complex and diverse tasks.
    \item We conducted a systematic evaluation of 10 open-source and 4 closed-source LLMs, revealing the strengths and weaknesses of different models in operational scheduling scenarios. The results serve as a crucial reference for LLM optimization, product development, and the practical deployment of LLMs in real-world telecommunications scenarios.
\end{itemize}

The key findings of the study are as follows:
\begin{itemize}
    \item Closed-source LLMs are not irreplaceable, as the open-source LLM DeepSeek-V3 demonstrated outstanding performance across all tasks, showcasing its strong potential in operational scheduling tasks. Moreover, in most tasks, the performance under the few-shot setting significantly surpassed that of the zero-shot setting, indicating that a modest number of examples can effectively enhance an LLM’s task adaptation and overall performance.
    \item The maturity ranking of operations scheduling tasks is as follows: report analysis tasks $\textgreater$ intent recognition tasks $\textgreater$ knowledge Q\&A tasks $\textgreater$ report generation tasks $\textgreater$ homogeneous ticket recommendation tasks.
    \item LLMs exhibit proficiency in descriptive text generation. In the report analysis task, a comprehensive statistical table was provided, allowing the LLM to focus solely on data analysis and the derivation of conclusions. With the highest score reaching 89.52, this result demonstrates the strong suitability of LLMs for applications involving data analysis and content generation.
    \item Conversely, LLMs are less proficient at performing statistical tasks. In report generation tasks, where LLMs were required to first compute statistics and then generate a textual report, the performance was suboptimal, highlighting the need for external tools to assist with data computation.
    \item Additionally, LLMs show limitations in handling multiple-choice questions with numerous options. Specifically, in the homogeneous ticket recommendation task, the redundancy of options resulted in a highest score of only 28.17. In practical applications, this necessitates ongoing LLM optimization and fine-tuning to enhance accuracy and stability for such tasks.
\end{itemize}

These contributions and findings provide critical insights into the practical application of LLMs within telecommunications operational scheduling, paving the way for further research and refined approaches in this domain.

\section{Related work}\label{sec2}

\subsection{General LLMs and benchmarks}\label{subsec2.1}

In recent years, the rapid development of LLMs, such as GPT-4o~\cite{achiam2023gpt}, GLM-4 ~\cite{glm2024chatglm}, Qwen-2.5~\cite{yang2024qwen2}, and DeepSeek-V3~\cite{liu2024deepseek}, has significantly propelled advancements in the field of NLP. These models, trained extensively on massive datasets, have demonstrated robust capabilities in language understanding and generation, achieving state-of-the-art performance across a wide range of tasks. As LLMs continue to evolve and find broader applications, the need for scientific and systematic evaluation of their performance has become a paramount concern for both academia and industry~\cite{guo2023evaluating, chang2024survey}.

Evaluation benchmarks for general-purpose LLMs can generally be categorized into two main types. The first category consists of comprehensive benchmarks that encompass a variety of natural language understanding and generation tasks. These benchmarks aim to assess LLMs’ adaptability in tasks such as text classification, reading comprehension, sentiment analysis, and text generation. Early examples of such benchmarks include GLUE~\cite{wang2018glue} and CLUE~\cite{xu2020clue}, with more recent developments exemplified by SuperGLUE~\cite{xu2023superclue} and A-Eval~\cite{lian2024best}. The second category comprises benchmarks structured around exam-style questions from various academic disciplines. These benchmarks prioritize the assessment of LLMs’ mastery of interdisciplinary knowledge and reasoning skills. Notable examples in this category include AGIEval~\cite{zhong2024agieval}, MMLU-Redux~\cite{wang2024mmlu}, C-Eval~\cite{huang2023c}, and SciKnowEval~\cite{feng2024sciknoweval}, which evaluate models through exam items covering diverse subjects and varying levels of difficulty. While these general benchmarks play a critical role in assessing the broad capabilities of large models, their lack of domain specificity limits their effectiveness in accurately measuring performance in specific industry contexts.

\subsection{Telecommunications LLMs and benchmarks}\label{subsec2}

In recent years, the application of LLMs has expanded across various vertical industries, including healthcare~\cite{santorinaios2024medguard, ye2025gmai, qiu2024towards}, finance~\cite{zhu2024benchmarking, krumdick2024bizbench, zhao2024knowledgefmath, guo2025flame}, law~\cite{dai2025laiw, yue2023disc}, and education~\cite{huang2023c}. However, research on the application of LLMs in the telecommunications sector remains relatively limited. Currently, the three major operators in the telecommunications industry have successively launched their self-developed, domain-specific LLMs. For instance, China Mobile introduced the “Jiutian Customer Service LLM”, which supports intelligent Q\&A and script generation in online customer service systems, enhancing service efficiency~\cite{wang2024smart}. China Unicom released the “Honghu Graphic LLM”, focusing on value-added services such as image generation and video editing, and offering customized solutions tailored to industry demands~\cite{yilma2025telecomrag}. Meanwhile, China Telecom unveiled the Telechat LLM~\cite{he2024telechat}, which aims to empower data middleware, intelligent customer service, and smart governance scenarios~\cite{zhou2024large}, thereby facilitating the intelligent upgrading of the communications industry. By leveraging the operators’ computational advantages and industry expertise, these large models provide important support for digital transformation.

The telecommunications domain primarily encompasses customer service, network management, and operations scheduling. While TeleEval-CS~\cite{li2024performance} has explored the evaluation of LLM capabilities in the customer service domain, operational scheduling aims to achieve comprehensive coordination and command scheduling of networks, services, risks, and personnel events. However, due to the complexity of the scenarios and the diversity of tasks, there is currently a lack of a universal evaluation system, making it challenging to comprehensively assess the actual capabilities and potential of LLMs in this field. Therefore, constructing an evaluation framework tailored to the operations scheduling domain in telecommunications is of paramount importance. To address this gap, our TeleEval-OS framework integrates real-world telecommunications operations scheduling scenarios and systematically organizes evaluation datasets from four perspectives: intelligent ticket creation, intelligent handling, intelligent ticket closure, and intelligent evaluation. In conjunction with specific task requirements, we designed basic NLP, knowledge Q\&A, report generation, and report analysis tasks across four levels to comprehensively evaluate the integrated capabilities of LLMs in the telecommunications operational scheduling domain.

\section{Benchmark construction}\label{sec3}

\begin{figure}[h]
\centering
\includegraphics[width=1\textwidth]{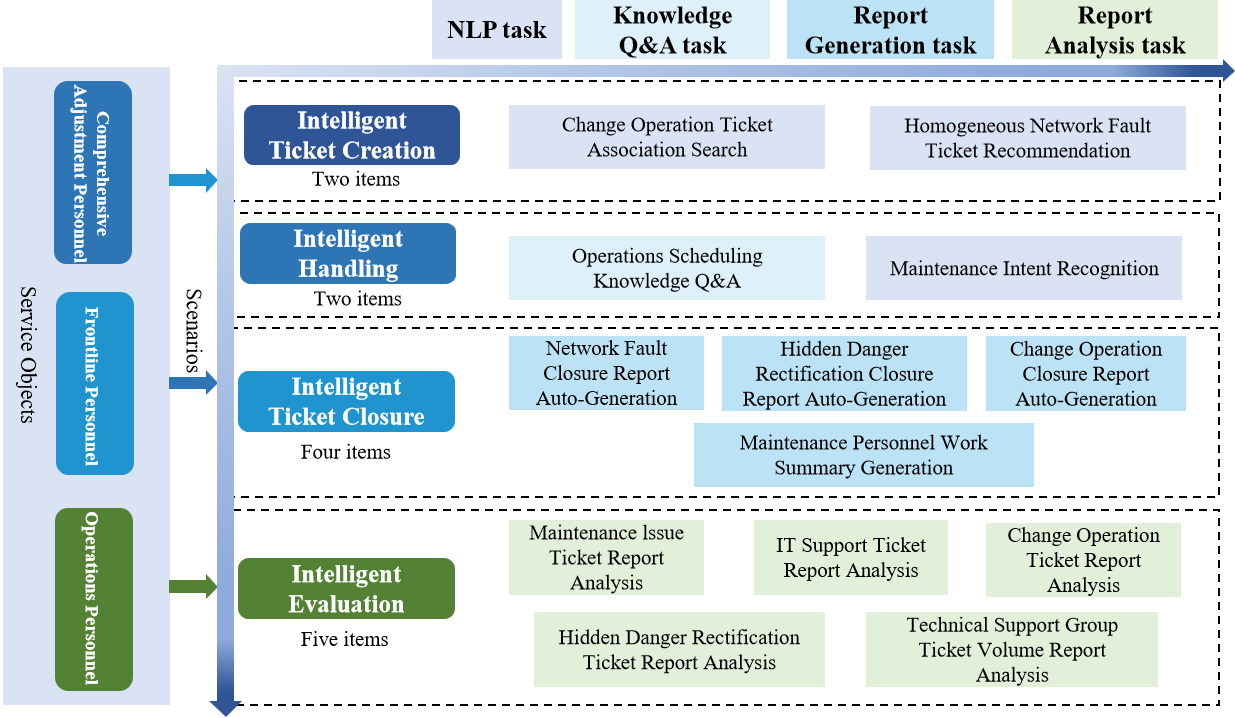}
\caption{Overview of the framework in telecommunications operations scheduling domain}\label{fig1}
\end{figure}

In collaboration with experts from the fields of operation scheduling, artificial intelligence, and related areas, we systematically categorized the telecommunications operational scheduling domain from two dimensions: scenarios and capabilities. Based on this, we established a scalable framework for multi-scenario and multi-task evaluation. As shown in Figure ~\ref{fig1}, the scenario dimension divides the domain into four key stages: intelligent ticket creation, intelligent handling, intelligent ticket closure, and intelligent evaluation. In terms of capabilities, the performance of LLMs in operation scheduling tasks is categorized into four hierarchical levels, arranged in order of increasing difficulty: basic NLP, knowledge Q\&A, report generation, and report analysis. These levels collectively encompass 13 subtasks, providing a comprehensive evaluation framework. This classification framework not only provides a structured approach for evaluating the performance of LLMs in the telecommunications operation scheduling domain but also establishes a foundation for future task expansion and optimization.

Specifically, the basic NLP tasks mainly involve simple classification tasks, including two homogeneous ticket recommendation tasks (change operation ticket association search and homogeneous network fault ticket recommendation) and an intent recognition task. The knowledge Q\&A tasks are designed to provide intelligent question answering within operation scheduling scenarios. The report generation category comprises three types of automatic ticket closure report generation tasks (network fault tickets, hidden danger rectification tickets, and change operation tickets) as well as a summary generation task for field maintenance personnel. Finally, the report analysis tasks are designed to generate structured reports based on ticket statistics, specifically covering analyses of tickets related to maintenance issues, IT support, change operations, hidden danger rectification, and technical support groups.

\subsection{Operation scheduling scenarios}\label{subsec3.1}

\subsubsection{Intelligent ticket creation scenario}\label{subsubsec3.1.1}

Intelligent ticket creation constitutes the initial phase of operation scheduling, with its core objective being the implementation of homogeneous ticket recommendation through basic NLP capabilities. Specifically, when creating a new ticket, the system automatically identifies and recommends historically similar tickets from the repository based on critical information such as ticket title and issue type. These recommended tickets exhibit semantic similarity in title keywords or categorical alignment in problem types. This functionality not only significantly enhances operational efficiency in ticket creation but also serves as reference material for subsequent ticket completion operations, thereby reducing redundant efforts and improving the accuracy of overall dispatch workflows.

\subsubsection{Intelligent handling creation scenario}\label{subsubsec3.1.2}

Intelligent handling is the core stage of operations scheduling, designed to provide intelligent support for operational personnel. This includes assisting installation and maintenance personal in household broadband installation and maintenance tasks, enabling them to efficiently complete work order processing.

\begin{itemize}
    \item Basic NLP Capability: This involves extracting implicit semantics from text inputs by installation and maintenance personnel to determine the true purpose and needs behind their statements and classify the questions they raise. For instance, during broadband installation, it can discern whether the technician is querying information on “installation card issues” or “MobiBox playback errors”. By automating intent recognition, this reduces the need for manual intervention, thereby accelerating the handling speed of installation and maintenance tasks.
    \item Knowledge Q\&A Capability: By retrieving information from a specialized knowledge base in the field of operations scheduling, relevant historical expertise can be quickly accessed. This enables intelligent question answering based on professional knowledge, thereby enhancing both the efficiency and accuracy of problem resolution.
\end{itemize}

\subsubsection{Intelligent ticket closure scenario}\label{subsubsec3.1.3}
Intelligent ticket closure represents the final stage of ticket processing, where LMMs primarily facilitate report generation by automating the summarization and feedback of the entire process and its outcomes. The intelligent ticket closure scenario encompasses two primary tasks: ticket closure report generation and maintenance personnel work summary generation. The ticket closure report generation task focuses on automatically summarizing and analyzing key information such as processing efficiency and remediation outcomes. Additionally, it provides insights and recommendations for process improvement. Meanwhile, the maintenance personnel work summary generation task is designed to automatically summarize and analyze the daily workload of maintenance staff, including the number of tickets processed, work types, and efficiency metrics. This task also delivers personalized work recommendations to help employees enhance their productivity

\subsubsection{Intelligent evaluation scenario}\label{subsubsec3.1.4}
The intelligent evaluation phase serves as the optimization component within ticket processing, where LLMs provide structured report analysis capabilities to enhance overall operational efficiency and service quality through data analytics and feedback mechanisms. This scenario includes five distinct ticket report analysis tasks, which leverage existing statistical data to analyze data distribution trends, extract key insights, and generate structured reports. Through automated analysis, the process significantly reduces the time required for manual data organization and summarization, thereby improving overall work efficiency.

\subsection{Construction of datasets for operational scheduling tasks}\label{subsec3.2}

\begin{figure}[h]
\centering
\includegraphics[width=1\textwidth]{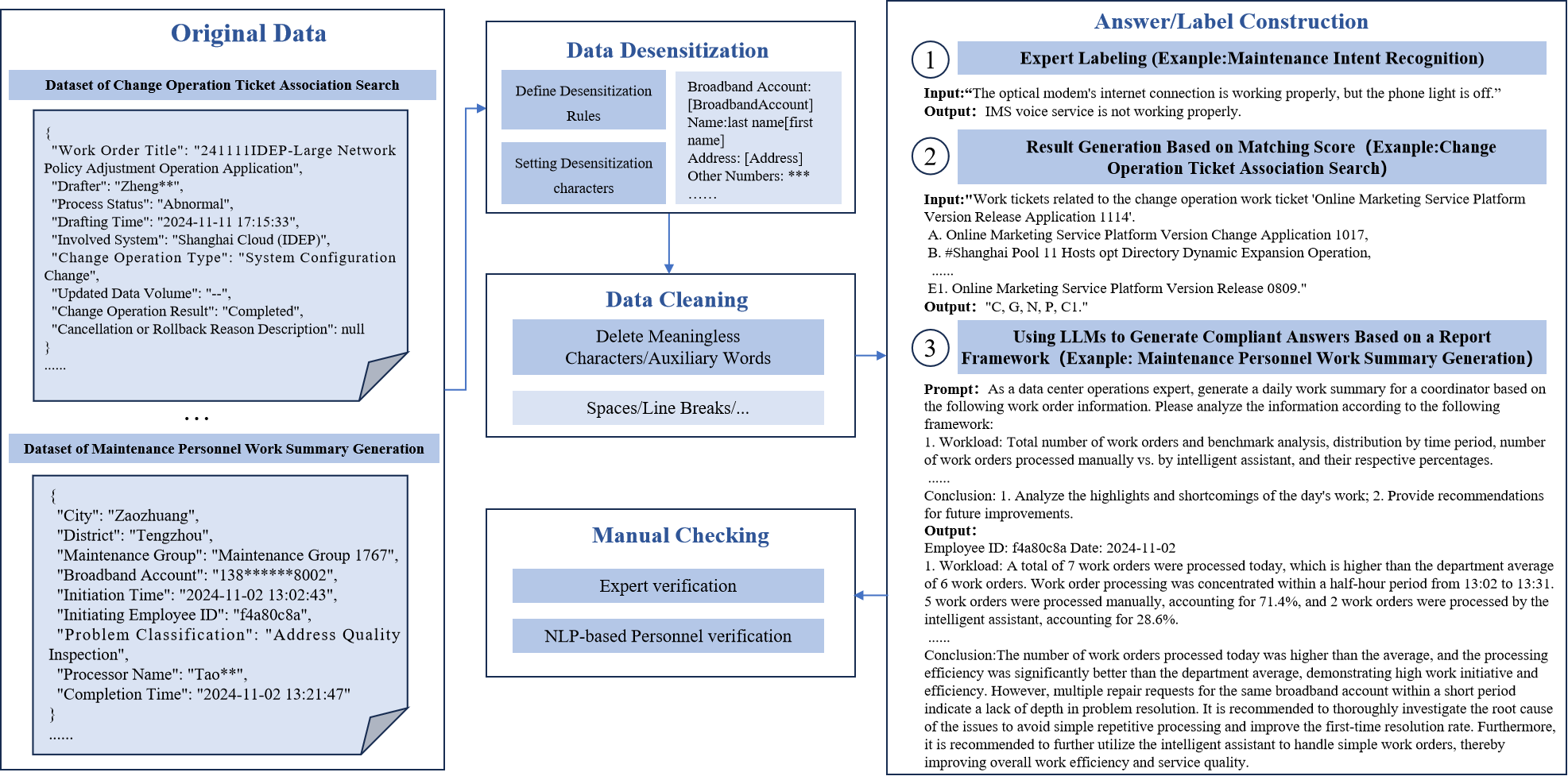}
\caption{Overview of the framework in telecommunications operations scheduling domain}\label{fig2}
\end{figure}

To evaluate the capabilities of LLMs in the domain of operational scheduling, we have constructed a proprietary dataset tailored to this field. The dataset construction process follows a unified workflow, as illustrated in Figure~\ref{fig2}, and includes steps such as data desensitization, data cleaning, answer or label generation, and data verification. First, to ensure customer privacy and data security, we apply regular expressions or hash functions to replace sensitive information, including broadband account numbers, maintenance personnel addresses, and names. Subsequently, meaningless characters, such as redundant spaces and special symbols, are removed from the anonymized text. Next, appropriate answers or structured tables are generated according to the characteristics of each dataset. The methods employed include: (a) manual annotation by business experts, (b) result generation based on matching score calculations, and (c) the use of LLMs to produce answers conforming to a prescribed report framework. Finally, to ensure data quality, a dual verification process is conducted by both business experts and NLP specialists, ensuring the accuracy and reliability of the dataset for downstream tasks. Specifically, we constructed 15 datasets across 13 subtasks, with a total of 10.4K samples. These datasets cover tasks in basic NLP, knowledge Q\&A, report generation, and report analysis, with detailed information provided in Table~\ref{tab1}.

\begin{table*}[h]
\centering
\caption{Overview of the evaluation dataset and metrics.}
\resizebox{\textwidth}{!}{ 
\begin{tabular}{@{}lllllll@{}}
\toprule
Scenes& Dataset& Task&Major Task Type& Minor Task Type&Size &Metrics\\
\midrule
\multirow{2}*{Intelligent Ticket Creation}
&COTAS&Change Operation Ticket Association Search&NLP task&Text Classification&500&F1\\ 
&HNFTR&Homogeneous Network Fault Ticket Recommendation&NLP task&Text Classification&151&F1\\
\midrule
\multirow{4}*{Intelligent Ticket Handling}
&OSEQA&Operations Scheduling Knowledge Q\&A&Knowledge Q\&A &Knowledge Q\&A&87&RougeL\\
&MIR-24&Maintenance Intent Recognition&NLP task&Text Classification&1653&Accuracy\\
&MIR-45&Maintenance Intent Recognition&NLP task&Text Classification&1637&Accuracy\\
&MIR-83&Maintenance Intent Recognition&NLP task&Text Classification&2000&Accuracy\\
\midrule
\multirow{4}*{Intelligent Ticket Closure}
&NFT-CRAG&Network Fault Closure Report Auto-Generation&Report Generation &Text Generation&100&RougeL\\
&HDRT-CRAG&Hidden Danger Rectification Closure Report Auto-Generation&Report Generation&Text Generation&100&RougeL\\
&COT-CRAG&Change Operation Closure Report Auto-Generation&Report Generation &Text Generation&100&RougeL\\
&MPWSG&Maintenance Personnel Work Summary Generation&Report Generation &Text Generation&100&RougeL\\
\midrule
\multirow{5}*{Intelligent Ticket Evaluation}
&MIT-RA&Maintenance Issue Ticket Report Analysis&Report Analysis&Text Generation&1500&Judger model\\
&ITST-RA&IT Support Ticket Report Analysis&Report Analysis&Text Generation&1661&Judger model\\
&TSGTV-RA&Technical Support Group Ticket Volume Report Analysis&Report Analysis &Text Generation&549&Judger model\\
&HDRT-RA&Hidden Danger Rectification Ticket Report Analysis&Report Analysis&Text Generation&43&Judger model\\
&COT-RA&Change Operation Ticket Report Analysis&Report Analysis&Text Generation&247&Judger model\\
\bottomrule
\end{tabular}}
\label{tab1}
\end{table*}

\section{Experiments}\label{sec4}

In this study, we systematically evaluated several state-of-the-art open-source and closed-source LLMs using the TeleEval-OS benchmark. Our primary objective was to investigate these models’ comprehensive capabilities across the intelligent ticket creation, intelligent handling, intelligent ticket closure and intelligent evaluation scenarios, with a particular focus on their performance in basic NLP, knowledge Q\&A, report generation, and report analysis tasks.

\subsection{Evaluation settings}\label{subsec4.1}

\subsubsection{Datasets and metrics}\label{subsubsec4.1.1}

The datasets utilized in this study are detailed in Table~\ref{tab1}, with each scenario containing multiple subtask datasets and their corresponding evaluation metrics. In the intelligent ticket creation scenario, we employ the F1 score to evaluate the performance of NLP tasks. The F1 score is a widely used metric for classification tasks, providing a balanced measure of model performance by computing the harmonic mean of precision and recall. For the intelligent ticket handling scenario, we use accuracy to assess the NLP task performance of LLMs. Accuracy measures the proportion of correctly predicted samples out of the total number of samples, offering an intuitive representation of classification performance. For both the knowledge Q\&A and report generation tasks, the Rouge-L metric is selected for evaluation. Rouge-L is a text similarity measure based on the longest common subsequence (LCS), which assesses the quality of generated content by computing the longest matching sequence between the predicted and reference texts. For the report analysis tasks, We utilize ChatGPT-4o as the judge model to perform a comprehensive assessment of the generated reports. The evaluation criteria include content completeness, data interpretation accuracy, analytical quality, formatting compliance, and clarity of expression.

\subsubsection{Experimental setup}\label{subsubsec4.1.2}

In this study, we employ both zero-shot and few-shot evaluation methods. Specifically, for the few-shot experiments on the intent recognition datasets (MIR-24, MIR-45, and MIR-83), we randomly select 20 samples from the validation set as contextual examples; for all other data sets, we randomly select 2 samples from the validation set to serve as contextual examples.

\subsubsection{Models}\label{subsubsec4.1.3}

This study evaluates a range of representative open-source and closed-source LLMs from both domestic and international sources, covering various model scales and architectures. The evaluation specifically focuses on their performance in Chinese language tasks. The closed-source LLMs include ChatGPT-4o\footnote{\url{ https://openai.com/index/hello-gpt-4o/}} by OpenAI, Claude-3.5-Sonnet\footnote{\url{ https://www.anthropic.com/news/claude-3-5-sonnet}} by Anthropic, Gemini-1.5-Pro\footnote{\url{https://deepmind.google/technologies/gemini/}} by Google, and GLM-4-Plus\footnote{\url{https://open.bigmodel.cn/}} by Zhipu AI, all of which represent the state-of-the-art in the field of LLMs. The open-source LLMs include internationally popular models such as Llama-3.2-3B-Instruct\footnote{\url{https://modelscope.cn/models/LLM-Research/Llama-3.2-3B-Instruct}}  and Gemma-2-9B-Instruct\footnote{\url{https://modelscope.cn/models/LLM-Research/Gemma-2-9B-Instruct}}, as well as a series of domestic open-source LLMs, including Qwen-2.5-7B-Instruct\footnote{\url{https://modelscope.cn/models/Qwen/Qwen-2.5-7B-Instructt}}, TeleChat-2-35B\footnote{\url{https://modelscope.cn/models/TeleAI/TeleChat2-35B}}, Yi-1.5-9B-Chat\footnote{\url{https://modelscope.cn/models/01ai/Yi-1.5-9B-Chat}}, and MiniCPM-3-4B\footnote{\url{https://modelscope.cn/models/OpenBMB/MiniCPM-3-4B}}. 

Table~\ref{tab2} presents details regarding the organizations behind these models, their sizes, and their respective access methods.

\begin{table}[h]
\caption{The LLMs evaluated in our work.}\label{tab2}%
\begin{tabular}{@{}llll@{}}
\toprule
Model& Developer& Size & Access \\
\midrule
ChatGPT-4o & OpenAI & undisclosed & API \\
Claude-3.5-Sonnet & Anthropic & undisclosed & API \\
GLM-4-Plus & Zhipu AI & undisclosed & API \\
Gemini-1.5-Pro & Google & undisclosed & API \\
\midrule
DeepSeek-V3 & DeepSeek & 671B & API \\
Qwen-2.5-7B-Instruct & Alibaba & 7B & Weights \\
Qwen-2.5-14B-Instruct & Alibaba & 14B & Weights \\
Qwen-2.5-72B-Instruct & Alibaba & 72B & API \\
TeleChat-2-35B & TeleAI & 35B & Weights \\
Llama-3.2-3B-Instruct & Meta & 3B & Weights \\
Yi-1.5-9B-Chat & Alibaba & 9B & Weights \\
GLM-4-9B-Chat & Zhipu AI & 9B & Weights \\
Gemma-2-9B-Instruct & Google & 9B & Weights \\
MiniCPM-3-4B & MiniCPM & 4B & Weights \\
\botrule
\end{tabular}
\end{table}

\subsection{Main results}\label{subsec4.2}

Table~\ref{tab3} presents the performance of 14 representative LLMs across various tasks in the telecommunications operation scheduling domain. First, the overall performance indicates that closed-source models generally exhibit superior results. Remarkably, GLM-4-Plus stands out with the highest average performance among these closed-source LLMs, closely followed by ChatGPT-4o and Claude-3.5-Sonnet. On the other hand, within the open-source category, DeepSeek-V3 achieves the best average performance, even surpassing all closed-source models, highlighting the potential and competitiveness of open-source LLMs in vertical domain applications.

\begin{table*}[h]
\centering
\caption{Performance of LLMs on TeleEval-OS benchmark under zero-shot and few-shot settings. Bold and underline represent the best and the second best respectively. The AVG column is the average score of each model across all tasks.}
\label{tab3}
\resizebox{\textwidth}{!}{
\begin{tabular}{@{\extracolsep{\fill}}lllllllll@{}}
\toprule
\multirow{2}{*}{Model} & \multicolumn{2}{c}{NLP Task} & \multicolumn{2}{c}{Knowledge Q\&A} & \multicolumn{2}{c}{Report Generation} & \multirow{2}{*}{\shortstack{Report\\Analysis}} & \multirow{2}{*}{AVG} \\
\cmidrule(lr){2-3} \cmidrule(lr){4-5} \cmidrule(lr){6-7}
 & 0-shot & 5-shot & 0-shot & 5-shot & 0-shot & 5-shot & & \\
\midrule

\multicolumn{9}{@{}c}{\textit{Open-source LLMs}} \\
GLM-4-Plus         & 50.43 & \textbf{54.81} & 61.58 & 68.03 & \textbf{48.13} & \textbf{58.26} & 86.20 & \textbf{60.49} \\
ChatGPT-4o         & \underline{51.03} & 53.48 & \underline{62.47} & \textbf{68.92} & 45.64 & \underline{55.93} & \textbf{89.52} & \underline{60.31} \\
Claude-3.5-Sonnet  & 50.10 & 54.03 & \textbf{63.62} & \underline{68.28} & \underline{46.77} & 53.27 & \underline{89.24} & 59.95 \\
Gemini-1.5-Pro     & 51.79 & \underline{54.14} & 55.10 & 64.79 & 44.97 & 55.35 & 88.36 & 59.70 \\

\addlinespace[0.3em]
\multicolumn{9}{@{}c}{\textit{Closed-source LLMs}} \\
DeepSeek-V3          & \textbf{51.45} & \textbf{55.42} & \textbf{64.00} & \underline{67.67} & \textbf{49.08} & \textbf{60.27} & \textbf{88.48} & \textbf{61.83} \\
Qwen-2.5-72B-Instruct& \underline{49.97} & \underline{51.89} & \underline{63.93} & \textbf{68.21} & \underline{45.07} & 55.23 & \underline{87.64} & \underline{59.23} \\
Qwen-2.5-14B-Instruct& 47.44 & 51.30 & 59.76 & 67.18 & 42.96 & 51.63 & 84.00 & 56.76 \\
GLM-4-9B-Chat        & 44.70 & 47.64 & 60.16 & 64.48 & 43.53 & \underline{56.02} & 78.44 & 55.07 \\
Gemma-2-9B-Instruct  & 44.93 & 47.94 & 50.69 & 55.96 & 43.58 & 54.51 & 81.36 & 54.81 \\
TeleChat-2-35B       & 40.68 & 42.40 & 60.83 & 65.96 & 44.41 & 54.58 & 85.40 & 54.60 \\
Qwen-2.5-7B-Instruct & 43.23 & 45.83 & 58.11 & 64.20 & 42.28 & 53.64 & 82.16 & 54.48 \\
MiniCPM-3-4B         & 35.78 & 41.56 & 51.57 & 62.32 & 43.76 & 53.61 & 76.68 & 50.94 \\
Yi-1.5-9B-Chat       & 35.39 & 38.99 & 58.69 & 63.23 & 40.53 & 54.67 & 79.68 & 50.92 \\
Llama-3.2-3B-Instruct& 23.02 & 36.22 & 41.02 & 51.40 & 39.14 & 53.66 & 60.74 & 42.54 \\
\bottomrule
\end{tabular}
}

\begin{minipage}{\textwidth}
\footnotesize
Note: All scores except report analysis have been standardized as percentages.
\end{minipage}
\end{table*}

Furthermore, regarding model parameter scale, there is a general correlation between model performance and parameter size, although performance varies considerably across different tasks. Large-scale LLMs, such as Qwen-2.5-72B-Instruct, exhibit consistently strong performance across all tasks, with an average score exceeding 59. While medium-scale LLMs show a slight decline in overall performance, they remain competitive in specific tasks, such as knowledge Q\&A. Small-scale LLMs, including MiniCPM-3-4B and Llama-3.2-3B-Instruct, perform relatively weakly in basic NLP tasks. However, their performance gap narrows in more complex tasks such as report generation, suggesting that the relationship between model size and performance is not strictly linear. It is worth noting that some optimized small-scale LLMs, such as MiniCPM-3-4B, achieve performance comparable to or even surpassing that of certain medium-scale models in specific tasks.

Lastly, the results of the few-shot experiments indicate that all models exhibit significant performance improvements across various telecommunications operation scheduling tasks, but the degree of improvement is closely related to task complexity and model size. From the perspective of task difficulty, in basic NLP tasks, where models already possess strong language understanding capabilities, the few-shot improvement is relatively limited. As task complexity increases, the performance gains achieved through few-shot learning become more pronounced. This demonstrates that examples provide more structured information for challenging tasks, leading to greater performance benefits. From the perspective of model size, large-scale LLMs (such as GLM-4-plus) demonstrate relatively stable performance gains under few-shot settings due to their strong foundational capabilities. In contrast, smaller-scale models exhibit more substantial performance improvements in few-shot settings, particularly in complex tasks. This highlights the capability of few-shot learning to effectively compensate for the shortcomings of small-scale LLMs in zero-shot scenarios.

\subsection{Results on NLP tasks}\label{subsec4.3}

The basic NLP tasks primarily involve simple classification tasks, including two types of homogeneous ticket recommendation tasks and intent recognition tasks. Under the zero-shot setting, the performance of LLMs on basic NLP task datasets is shown in Table~\ref{tab4}. Overall, both the closed-source LLM Gemini-1.5-Pro and the open-source LLM DeepSeek-V3 achieve the highest average performances, both achieving average scores above 51. From the perspective of datasets, the scores for the homogeneous ticket recommendation datasets COTAS and HNFTR are generally low. For example, ChatGPT-4o scores only 20.77 in COTAS and 31.56 in HNFTR. These two datasets belong to the homogeneous ticket recommendation tasks, which use a multiple-choice format with a large number of options (an average of 109 options for COTAS and 96 options for HNFTR), leading to a significant drop in accuracy. Moreover, in the intent recognition task, the number of categories has a significant impact on model performance. For example, Gemini-1.5-Pro scores 76.13 on MIR-24, while its performance drops to 61.40 on MIR-83. This phenomenon indicates that as the number of categories increases, the task of discriminating among them becomes more complex for LLMs, which are consequently more prone to confusion and a decline in accuracy. However, smaller-scale models such as MiniCPM-3-4B and Yi-1.5-9B-Chat consistently score significantly lower than larger models across all tasks, and their weaknesses in performance are particularly evident in tasks with many options or categories. This reflects a significant limitation in the ability of small-scale LLMs to handle complex reasoning and large-scale category selection challenges.

\begin{table}[h]
\caption{LLMs performance in operations scheduling NLP tasks at zero-shot settings. Bold and underline represent the best and the second best respectively.}\label{tab4}
\begin{tabular*}{\textwidth}{@{\extracolsep\fill}lcccccc@{}}
\toprule%
Model & COTAS & HNFTR & MIR-24 & MIR-45 & MIR-83 & AVG \\
\midrule
\multicolumn{7}{@{}c}{\textit{Open-source LLMs}} \\
Gemini-1.5-Pro & \underline{23.35}  & \underline{31.43}  & \underline{76.13}  & \textbf{66.66}  & \textbf{61.40}  & \textbf{51.79}  \\ 
ChatGPT-4o & 20.77  & \textbf{31.56}  & \textbf{76.38}  & \underline{66.35}  & 60.10  & \underline{51.03}  \\
GLM-4-Plus & 21.71  & 31.21  & 73.51  & 65.52  & 60.20  & 50.43  \\
Claude-3.5-Sonnet & \textbf{24.09} & 29.49  & 75.51  & 62.49  & 58.90  & 50.10  \\ 

\addlinespace[0.3em]
\multicolumn{7}{@{}c}{\textit{Closed-source LLMs}} \\
DeepSeek-V3          & \textbf{23.68} & \textbf{31.26} & 76.32 & \textbf{65.40} & \textbf{60.57} & \textbf{51.45} \\
Qwen-2.5-72B-Instruct& \underline{20.07} & \underline{30.80} & \textbf{77.38} & \underline{64.32} & 57.28 & \underline{49.97} \\
Qwen-2.5-14B-Instruct& 16.91 & 24.12 & 74.89 & 63.88 & \underline{57.38} & 47.44 \\
Gemma-2-9B-Instruct  & 16.02 & 23.87 & 71.45 & 61.66 & 51.64 & 44.93 \\
GLM-4-9B-Chat        & 16.83 & 28.87 & 72.01 & 56.10 & 49.71 & 44.70 \\
Qwen-2.5-7B-Instruct & 13.64 & 21.76 & 70.76 & 58.31 & 51.69 & 43.23 \\
TeleChat-2-35B       & 13.78 & 14.62 & 67.83 & 57.36 & 49.81 & 40.68 \\
MiniCPM-3-4B         & 13.32 & 20.17 & 63.77 & 44.52 & 37.12 & 35.78 \\
Yi-1.5-9B-Chat       & 10.95 & 12.72 & 59.33 & 50.03 & 43.91 & 35.39 \\
Llama-3.2-3B-Instruct& 9.40  & 15.15 & 37.41 & 27.39 & 25.74 & 23.02 \\ 
\botrule
\end{tabular*}
\end{table}

\begin{figure}[h]
\centering
\includegraphics[width=1\textwidth]{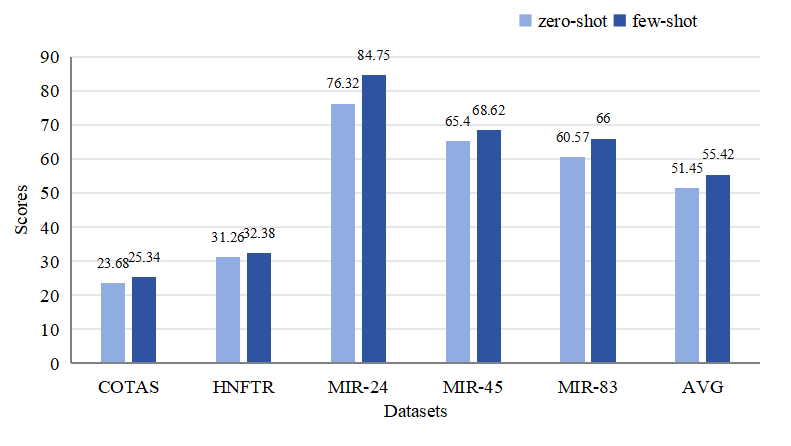}
\caption{The zero-shot and few-shot performance of DeepSeek-V3 on different datasets in NLP tasks.}\label{fig3}
\end{figure}

Under the few-shot setting, the performance of LLMs across different datasets for basic NLP tasks is presented in Table~\ref{tab5}. It can be observed that the overall performance trends of the LLMs under few-shot conditions are largely consistent with those observed in the zero-shot setting, though some differences exist. Overall, the few-shot setting significantly enhances the performance of LLMs on these tasks, as exemplified by Llama-3.2-3B-Instruct, whose average score increased from 23.02 in the zero-shot setting to 36.22 in the few-shot setting. As illustrated in Figure~\ref{fig3}, the intent recognition task shows particularly significant improvements with the incorporation of few-shot examples, with DeepSeek-V3 achieving an improvement on MIR-24 from 76.32 to 84.75. Conversely, in the ticket recommendation tasks, the improvements from few-shot examples are relatively modest. For instance, DeepSeek-V3 achieves only a slight improvement on COTAS, rising from 23.68 to 25.34. Notably, the performance of some smaller-scale LLMs declined under the few-shot setting. This degradation may be attributed to the increased context length resulting from the inclusion of additional examples, which can negatively impact the models’ comprehension abilities. In summary, while few-shot learning significantly improves the intent recognition task, the increase in the number of options and categories remains an important factor influencing overall model performance.

\begin{table}[h]
\caption{LLMs performance in operations scheduling NLP tasks at few-shot settings. Bold and underline represent the best and the second best respectively.}\label{tab5}
\begin{tabular*}{\textwidth}{@{\extracolsep\fill}lcccccc@{}}
\toprule%
Model & COTAS & HNFTR & MIR-24 & MIR-45 & MIR-83 & AVG \\
\midrule
\multicolumn{7}{@{}c}{\textit{Open-source LLMs}} \\
GLM-4-Plus        & \textbf{24.32} & \textbf{34.93} & 83.07 & 67.29 & 64.43 & \textbf{54.81} \\
Gemini-1.5-Pro    & 23.59 & 28.05 & \textbf{85.00} & 68.56 & \textbf{65.48} & \underline{54.14} \\
Claude-3.5-Sonnet & \underline{24.19} & 28.20 & 84.25 & \textbf{69.00} & \underline{64.49} & 54.03 \\
ChatGPT-4o        & 20.88 & \underline{30.05} & \underline{84.44} & \underline{68.75} & 63.28 & 53.48 \\ 

\addlinespace[0.3em]
\multicolumn{7}{@{}c}{\textit{Closed-source LLMs}} \\
DeepSeek-V3          & \textbf{25.34} & \textbf{32.38} & \underline{84.75} & \textbf{68.62} & \textbf{66.00} & \textbf{55.42} \\
Qwen-2.5-72B-Instruct& \underline{19.94} & \underline{29.18} & \textbf{85.82} & 64.38 & 60.15 & \underline{51.89} \\
Qwen-2.5-14B-Instruct& 19.00 & 27.20 & 83.19 & \underline{66.60} & \underline{60.52} & 51.30 \\
Gemma-2-9B-Instruct  & 17.47 & 18.40 & 80.88 & 64.70 & 58.27 & 47.94 \\
GLM-4-9B-Chat        & 18.10 & 28.30 & 82.63 & 57.49 & 51.69 & 47.64 \\
Qwen-2.5-7B-Instruct & 15.09 & 24.08 & 78.88 & 58.69 & 52.42 & 45.83 \\
TeleChat-2-35B       & 13.32 & 7.32  & 77.45 & 61.79 & 52.11 & 42.40 \\
MiniCPM-3-4B         & 13.01 & 13.77 & 75.95 & 56.67 & 48.40 & 41.56 \\
Yi-1.5-9B-Chat       & 9.62  & 7.75  & 74.20 & 56.35 & 47.04 & 38.99 \\
Llama-3.2-3B-Instruct& 7.53  & 11.25 & 71.01 & 50.85 & 40.46 & 36.22 \\
\botrule
\end{tabular*}
\end{table}

\subsection{Results on report generation tasks}\label{subsec4.4}

The report generation tasks focus on automatically summarizing and analyzing the key information from work tickets. These tasks include generating three types of ticket closure reports as well as work summaries for maintenance personnel. Under the zero-shot setting, Table~\ref{tab6} shows the performance of various LLMs on different report generation datasets. The open-source LLM DeepSeek-V3 achieved the highest average score of 49.08, outperforming all closed-source models, including the closed-source GLM-4-Plus, which followed closely with 48.13. This indicates that some open-source LLMs are competitive with or even slightly superior to closed-source models in report generation. From a dataset perspective, LLMs performed better on the MPWSG dataset, indicating that LLMs are more effective with this type of data. In contrast, all models scored relatively low on the COT-CRAG dataset, highlighting the challenges of this dataset. Notably, the open-source mid-scale model Telechat2-35B achieved performance comparable to closed-source models on datasets like HDRT-CRAG and COT-CRAG, highlighting the potential of mid-scale open-source LLMs for report generation tasks.

\begin{table}[h]
\caption{LLMs performance in operations scheduling report generation tasks at zero-shot settings. Bold and underline represent the best and the second best respectively.}\label{tab6}
\begin{tabular*}{\textwidth}{@{\extracolsep\fill}lccccc@{}}
\toprule%
Model & NFT-CRAG & HDRT-CRAG & COT-CRAG & MPWSG & AVG \\
\midrule
\multicolumn{6}{@{}c}{\textit{Open-source LLMs}} \\
GLM-4-Plus & \textbf{51.33} & \underline{48.18} & 42.54 & \textbf{50.45} & \textbf{48.13} \\ 
Claude-3.5-Sonnet & \underline{48.94} & 45.23 & \textbf{43.16} & \underline{49.74} & \underline{46.77} \\ 
ChatGPT-4o & 46.62 & \textbf{44.97} & \underline{42.77} & 48.19 & 45.64 \\ 
Gemini-1.5-Pro & 44.54 & 43.76 & 42.31 & 49.25 & 44.97 \\

\addlinespace[0.3em]
\multicolumn{6}{@{}c}{\textit{Closed-source LLMs}} \\
DeepSeek-V3 & \textbf{47.70} & \textbf{46.94} & \textbf{47.21} & \textbf{54.48} & \textbf{49.08} \\ 
Qwen-2.5-72B-Instruct & \underline{45.90} & 42.70 & \underline{43.06} & \underline{48.62} & \underline{45.07} \\ 
TeleChat-2-35B & 45.06 & \underline{44.67} & 42.81 & 45.08 & 44.41 \\ 
MiniCPM-3-4B & 45.81 & 42.78 & 41.14 & 45.31 & 43.76 \\ 
Gemma-2-9B-Instruct & 45.43 & 43.16 & 38.04 & 47.69 & 43.58 \\ 
GLM-4-9B-Chat & 41.78 & 43.26 & 42.08 & 47.01 & 43.53 \\ 
Qwen-2.5-14B-Instruct & 43.39 & 41.11 & 41.69 & 45.64 & 42.96 \\ 
Qwen-2.5-7B-Instruct & 39.62 & 42.04 & 42.20 & 45.26 & 42.28 \\ 
Yi-1.5-9B-Chat & 42.33 & 40.23 & 42.04 & 37.52 & 40.53 \\ 
Llama-3.2-3B-Instruct & 42.27 & 37.92 & 36.96 & 39.41 & 39.14 \\ 
\botrule
\end{tabular*}
\end{table}

\begin{figure}[h]
\centering
\includegraphics[width=1\textwidth]{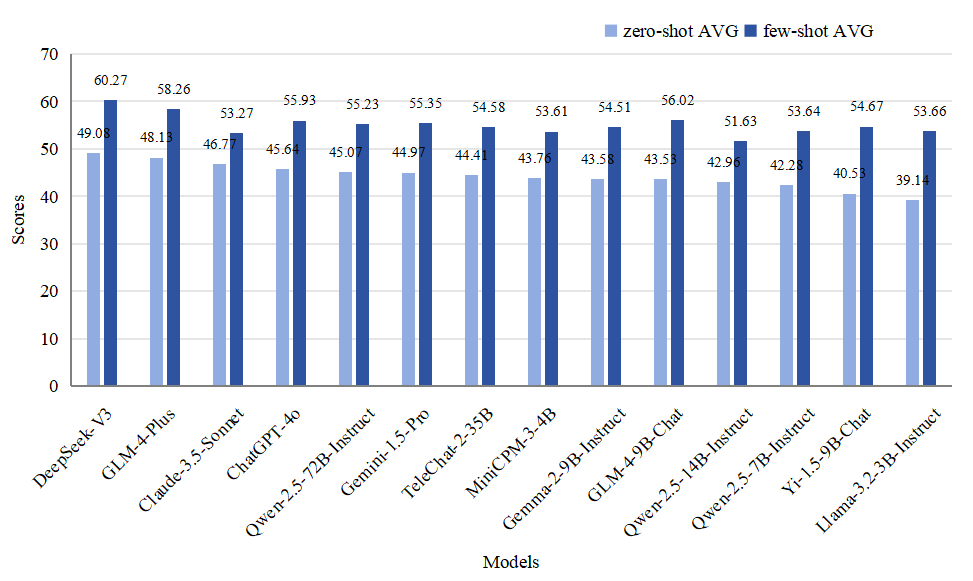}
\caption{The average zero-shot and few-shot performance of various LLMs on report generation tasks.}\label{fig4}
\end{figure}

Under the few-shot setting, Table~\ref{tab7} shows the performance of various LLMs on the report generation tasks across different datasets, while Figure~\ref{fig4} depicts the zero-shot and few-shot average scores for each LLM. The results reveal that the few-shot setting significantly enhances the performance of LLMs on report generation tasks. Notably, DeepSeek-V3 achieves the highest improvement, with its average score increasing from 49.08 to 60.27, outperforming all other LLMs. This suggests that providing examples helps identify key points in report results and content expression, thereby improving text generation quality. Additionally, smaller-scale models, such as GLM-4-9B-Chat and Yi-1.5-9B-Chat, show significant rank improvements under the few-shot setting, indicating that few-shot learning is particularly beneficial to these models. However, it is interesting to note that Qwen-2.5-14B-Instruct achieves a lower average score in the few-shot setting compared to Qwen-2.5-7B-Instruct. This counterintuitive result suggests that model performance is not determined solely by scale. Factors such as optimization strategies and task adaptability also play a critical role.

\begin{table}[h]
\caption{LLMs performance in operations scheduling report generation tasks at few-shot settings. Bold and underline represent the best and the second best respectively.}\label{tab7}
\begin{tabular*}{\textwidth}{@{\extracolsep\fill}lccccc@{}}
\toprule%
Model & NFT-CRAG & HDRT-CRAG & COT-CRAG & MPWSG & AVG \\
\midrule
\multicolumn{6}{@{}c}{\textit{Open-source LLMs}} \\
GLM-4-Plus & \textbf{62.64} & \textbf{53.68} & \textbf{53.57} & \textbf{63.14} & \textbf{58.26} \\ 
ChatGPT-4o & \underline{60.00} & 50.60 & \underline{50.59} & \underline{62.54} & \underline{55.93} \\ 
Gemini-1.5-Pro & 59.89 & \underline{51.72} & 49.66 & 60.13 & 55.35 \\ 
Claude-3.5-Sonnet & 56.43 & 50.61 & 46.75 & 59.29 & 53.27 \\ 

\addlinespace[0.3em]
\multicolumn{6}{@{}c}{\textit{Closed-source LLMs}} \\
DeepSeek-V3 & \textbf{64.44} & \textbf{55.94} & \textbf{54.85} & \textbf{65.83} & \textbf{60.27} \\ 
GLM-4-9B-Chat & 59.21 & \underline{52.49} & \underline{51.90} & 60.46 & \underline{56.02} \\ 
Qwen-2.5-72B-Instruct & 59.42 & 49.15 & 50.50 & \underline{61.85} & 55.23 \\ 
Yi-1.5-9B-Chat & 60.18 & 49.83 & 50.29 & 58.39 & 54.67 \\ 
TeleChat-2-35B & \underline{61.26} & 48.85 & 49.76 & 58.43 & 54.58 \\ 
Gemma-2-9B-Instruct & 55.18 & 51.26 & 51.02 & 60.56 & 54.51 \\ 
Llama-3.2-3B-Instruct & 59.10 & 49.70 & 48.61 & 57.22 & 53.66 \\ 
Qwen-2.5-7B-Instruct & 57.96 & 48.92 & 49.77 & 57.89 & 53.64 \\ 
MiniCPM-3-4B & 56.42 & 50.32 & 49.55 & 58.13 & 53.61 \\ 
Qwen-2.5-14B-Instruct & 55.25 & 47.54 & 46.41 & 57.31 & 51.63 \\
\botrule
\end{tabular*}
\end{table}

\subsection{Knowledge Q\&A task results}\label{subsec4.5}

\begin{figure}[h]
\centering
\includegraphics[width=1\textwidth]{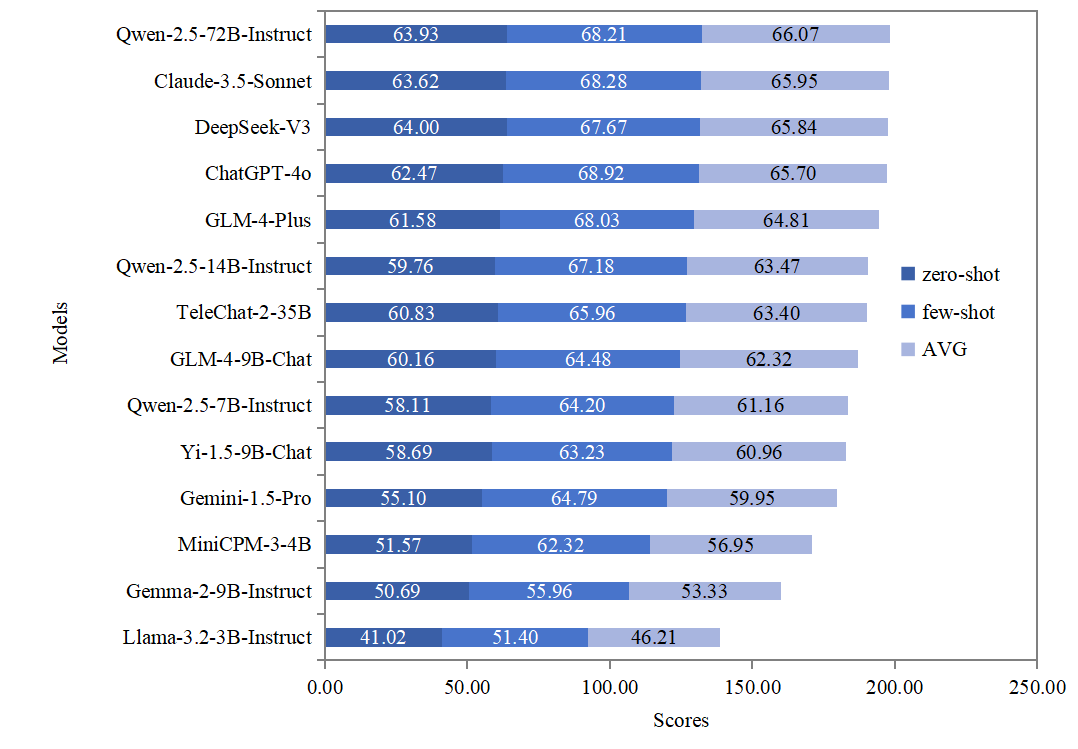}
\caption{LLMs performance in operations scheduling knowledge Q\&A tasks. Bold and underline represent the best and the second best respectively.}\label{fig5}
\end{figure}

The knowledge Q\&A task is designed to enable LLMs to answer user questions by referencing relevant documents. Its performance has been evaluated under both zero-shot and few-shot settings, as shown in Figure~\ref{fig5}. The experimental results show that Qwen-2.5-72B-Instruct achieves the best performance in this task with an average RougeL score of 66.07. This result even exceeds those of all closed-source LLMs and demonstrates exceptional question answering capabilities. In addition, under the few-shot setting, all models show significant performance improvements with average increases ranging from 5 to 10 points. These findings indicate that providing examples significantly improves the accuracy of answers generated by models.

\subsection{Report analysis task results}\label{subsec4.6}

The report analysis task is designed to generate structured reports from ticket statistical data. This task involves analyzing reports for five types of tickets. In our experiments, we adopt the state-of-the-art LLM model GPT-4o as the evaluation benchmark. Reports generated by various LLMs are evaluated comprehensively based on report content completeness, data interpretation accuracy, analysis quality, format compliance, and expression. Table~\ref{tab8} shows the performance of different LLMs on various report analysis datasets, with each score being the average of five evaluations. Overall, all LLMs exhibit strong performance in this task. Their performance demonstrates significant strengths in interpreting data distribution trends and conducting text analysis. 

Among the evaluated models, ChatGPT-4o stands out with an average score of 89.52, particularly achieving high scores of 92.40 and 91.00 on the TSGTV-RA and COT-RA datasets, respectively. These results highlight its exceptional performance and stability across multiple evaluation dimensions. Gemini-1.5-Pro and DeepSeek-V3 rank closely behind with average scores of 88.36 and 88.48. It is worth noting that there are variations in performance across datasets. For example, Qwen-2.5-72B-Instruct achieved a peak score of 93.40 on the ITST-RA dataset, while its performance on other datasets was relatively average. In addition, there is a strong correlation between model size and performance. Larger-scale LLMs with over 70 billion parameters generally outperform middle-sized and small-scale models. This difference may be due to the enhanced reasoning capabilities of larger models, which enable them to better understand report structures, perform data calculations, and accurately identify key trends.

\begin{table}[h]
\caption{LLMs performance in operations scheduling report analysis tasks. Bold and underline represent the best and the second best respectively.}\label{tab8}
\begin{tabular*}{\textwidth}{@{\extracolsep\fill}lcccccc@{}}
\toprule%
Model & MIT-RA & ITST-RA & TSGTV-RA & HDRT-RA & COT-RA & AVG \\
\midrule
\multicolumn{7}{@{}c}{\textit{Open-source LLMs}} \\
ChatGPT-4o & \underline{87.00} & 88.60 & \textbf{92.40} & \textbf{88.60} & 91.00 & \textbf{89.52} \\ 
Claude-3.5-Sonnet & \textbf{89.40} & \underline{89.40} & 86.00 & \textbf{88.60} & \underline{92.80} & \underline{89.24} \\ 
Gemini-1.5-Pro & 83.60 & \textbf{91.40} & 87.20 & \underline{85.60} & \textbf{94.00} & 88.36 \\ 
GLM-4-Plus & 84.40 & 88.80 & \underline{88.00} & 80.20 & 89.60 & 86.20 \\  

\addlinespace[0.3em]
\multicolumn{7}{@{}c}{\textit{Closed-source LLMs}} \\
DeepSeek-V3 & \underline{86.60} & \underline{90.40} & \textbf{88.60} & \underline{86.00} & \textbf{90.80} & \textbf{88.48} \\ 
Qwen-2.5-72B-Instruct & \textbf{86.80} & \textbf{93.40} & 85.00 & \textbf{86.40} & 86.60 & \underline{87.64} \\ 
TeleChat-2-35B & 82.40 & 82.80 & \textbf{88.60} & 82.80 & \underline{90.40} & 85.40 \\ 
Qwen-2.5-14B-Instruct & 78.80 & 83.00 & \underline{86.00} & 83.60 & 88.60 & 84.00 \\ 
Qwen-2.5-7B-Instruct & 82.00 & 82.20 & 82.60 & 81.40 & 82.60 & 82.16 \\ 
Gemma-2-9B-Instruct & 81.40 & 77.80 & 81.40 & 80.00 & 86.20 & 81.36 \\ 
Yi-1.5-9B-Chat & 78.20 & 82.60 & 79.40 & 78.60 & 79.60 & 79.68 \\ 
GLM-4-9B-Chat & 79.20 & 80.60 & 80.80 & 78.80 & 72.80 & 78.44 \\ 
MiniCPM-3-4B & 82.20 & 77.40 & 74.60 & 68.40 & 80.80 & 76.68 \\ 
Llama-3.2-3B-Instruct & 49.60 & 61.30 & 54.00 & 67.60 & 71.20 & 60.74 \\
\botrule
\end{tabular*}
\end{table}

\subsection{Additional analysis}\label{subsec4.7}

Since the F1 scores for LLMs on the COTAS and HNFTR datasets are generally low, we introduce two additional metrics, specifically subset rate and intersection rate, to further evaluate model performance. The subset rate measures the probability that the predicted answer is completely contained within the correct answer, thereby reflecting prediction precision. In contrast, the intersection rate measures the likelihood of any overlap between the correct answer and the predicted answer and indicates coverage capability. As shown in Table~\ref{tab9}, the redundancy of answer options within the homogeneous ticket recommendation task has a significant impact on the performance of the LLMs. In both the COTAS and HNFTR datasets, the intersection rate for all LLMs is significantly higher than the corresponding F1 scores, reaching as high as 92.6\% for COTAS and 95.36\% for HNFTR. This discrepancy suggests that while the LLMs generally capture the correct answers, they are also prone to overgeneralization. Conversely, the subset rates are relatively low, with a maximum of 15.4\% for COTAS and 45.69\% for HNFTR, indicating a challenge for the models in precisely identifying the minimal set of correct answers. For instance, GLM-4-9B-Chat achieves an intersection rate of 92.6\% on the COTAS dataset, while its F1 score is only 16.83\%. This further demonstrates that although the model successfully identifies many correct selections, it also makes a significant number of incorrect predictions. In summary, the results show that within the homogeneous ticket recommendation task, the models often capture the correct answers; however, they also output a significant number of incorrect responses, which implies that improving the precision of identifying target answers remains a priority.

\begin{table}[h]
\caption{Evaluation of F1-score, subset matching, and intersection matching metrics for LLMs on cotas and hnftr datasets at zero-shot settings. Bold and underline represent the best and the second best respectively.}\label{tab9}
\begin{tabular*}{\textwidth}{@{\extracolsep\fill}lccccccc@{}}
\toprule%

\multirow{2}{*}{Model} & \multicolumn{3}{c}{COTAS} & \multicolumn{3}{c}{HNFTR} & \multirow{2}{*}{AVG} \\
\cmidrule(lr){2-4} \cmidrule(lr){5-7}
 & F1 & Subset & Intersection & F1 & Subset & Intersection & \\

\midrule
\multicolumn{8}{@{}c}{\textit{Open-source LLMs}} \\
GLM-4-Plus & 21.71 & 13.20 & \textbf{67.40} & 31.21 & 35.09 & \textbf{88.74} & \textbf{42.89} \\ 
Claude-3.5-Sonnet & \textbf{24.09} & \underline{13.80} & \underline{57.60} & 29.49 & 43.04 & \underline{85.43} & \underline{42.24} \\ 
Gemini-1.5-Pro & \underline{23.35} & 11.80 & 57.00 & \underline{31.43} & \textbf{45.69} & 83.44 & 42.12 \\ 
ChatGPT-4o & 20.77 & \textbf{15.40} & 49.00 & \textbf{31.56} & \underline{43.70} & \underline{85.43} & 40.98 \\   

\addlinespace[0.3em]
\multicolumn{8}{@{}c}{\textit{Closed-source LLMs}} \\
DeepSeek-V3 & \textbf{23.68} & 11.00 & 64.60 & \textbf{31.26} & \textbf{41.05} & 82.10 & \textbf{42.28} \\ 
GLM-4-9B-Chat & 16.83 & 2.00 & \textbf{92.60} & 28.87 & 6.62 & \textbf{95.36} & \underline{40.38} \\ 
Qwen-2.5-72B-Instruct & \underline{20.07} & 6.40 & 70.00 & \underline{30.80} & \underline{19.20} & \underline{92.71} & 39.86 \\ 
Qwen-2.5-14B-Instruct & 16.91 & 2.20 & \underline{79.80} & 24.12 & 3.97 & \underline{92.71} & 36.62 \\ 
MiniCPM-3-4B & 13.32 & \underline{11.40} & 70.60 & 20.17 & 13.24 & 78.80 & 34.59 \\ 
Gemma-2-9B-Instruct & 16.02 & 4.80 & 66.80 & 23.87 & 17.21 & 72.84 & 33.59 \\ 
Qwen-2.5-7B-Instruct & 13.64 & 5.20 & 61.80 & 21.76 & 0.66 & 84.10 & 31.19 \\ 
TeleChat-2-35B & 13.78 & 5.20 & 61.80 & 14.62 & 5.96 & 82.78 & 30.39 \\ 
Llama-3.2-3B-Instruct & 9.40 & \textbf{14.60} & 35.20 & 15.15 & 3.97 & 78.80 & 26.19 \\ 
Yi-1.5-9B-Chat & 10.95 & \underline{11.40} & 32.60 & 12.72 & 9.27 & 49.00 & 20.99 \\ 
\botrule
\end{tabular*}
\end{table}

\subsection{Discussion}\label{subsec4.8}

This study reveals the potential applications and limitations of LLMs in the field of operational scheduling. Experimental data demonstrate that DeepSeek-V3 achieves the optimal domain adaptability with a comprehensive score of 61.83. However, the performance of various LLMs shows complementary characteristics across different scenarios. Thus, selecting the optimal model based on task-specific characteristics is advised in practical applications. For intelligent ticket creation and intelligent ticket closure scenarios, DeepSeek-V3 is recommended due to its exceptional performance in text matching and generation. In intelligent handling scenarios, Claude-3.5-Sonnet is preferred for intent recognition tasks because of its strong capabilities in semantic understanding. For knowledge Q\&A tasks, Qwen-2.5-72B-Instruct is recommended due to its excellent performance in information extraction and knowledge integration. In intelligent evaluation scenarios, ChatGPT-4o demonstrates high suitability for data analysis and report generation, making it the LLM of choice.

From the perspective of task maturity, the report analysis task demonstrates a higher level of maturity due to the provision of comprehensive data tables, with the highest score reaching 89.52. Intent recognition and knowledge Q\&A tasks also perform relatively stably, with highest scores of 70.54 and 66.07, respectively. In contrast, report generation tasks, where LLMs first summarize data and then generate textual reports, exhibit weaker performance, with a maximum score of only 54.67. This suggests that incorporating professional statistical tools to assist LLMs in data processing may improve outcomes in the future. For homogeneous ticket recommendation tasks, the redundancy of options leads to a maximum score of only 28.17, indicating the need for ongoing optimization and fine-tuning to enhance accuracy and stability in practical applications.

In summary, when deploying operational scheduling systems, multi-LLM collaboration or scenario-specific customization can be adopted based on actual business needs. Selecting the most suitable model and continuously optimizing and fine-tuning based on the maturity of different tasks can further improve overall scheduling efficiency and decision-making quality.

\section{Conclusion}\label{sec5}

This study introduces the first specialized evaluation benchmark in the field of telecommunications operational scheduling, named TeleEval-OS. The benchmark systematically assesses the comprehensive capabilities of LLMs across the entire telecommunication operational scheduling process through a four-phase evaluation framework: intelligent ticket creation, intelligent handling, intelligent ticket closure, and intelligent evaluation. TeleEval-OS provides a scientific foundation for model selection in practical scenarios. The evaluation results demonstrate that open-source models can achieve performance levels comparable to commercial closed-source models, with DeepSeek-V3 exhibiting the best overall performance across all tasks within TeleEval-OS. In future work, we aim to expand the TeleEval-OS benchmark to include a broader range of telecommunication operational scheduling scenarios. Additionally, we will explore techniques such as model fine-tuning and multimodal integration to further enhance the performance and generalization capabilities of LLMs in this domain.

\bmhead{Author Contributions}
Development of the evaluation framework: Yanyan Wang and Fei Li; data collection and organization: Yin Xu, Long Zhao and Kuang Xu; large model evaluation: Junli Liang, Yunlong Liu, Yiming Xu, Zhengwang Jiang and Zhehe Li; paper writing: Yingying Wang; paper review: Qi Song and Xiangyang Li. All authors provided comments on previous versions of the manuscript. All authors have read and approved the final manuscript.

\bmhead{Funding}
The research is funded by the Anhui Postdoctoral Scientific Research Program Foundation (No. 2024A768). The research is partially supported by the National Key R\&D Program of China under Grant No. 2021ZD0110400, the Innovation Program for Quantum Science and Technology (No. 2021ZD0302900), the China National Natural Science Foundation (Grant No. 62132018), the "Pioneer" and "Leading Goose" R\&D Program of Zhejiang (No. 2023C01029), and the "Leading Goose" R\&D Program of Zhejiang (No. 2023C01143).

\bmhead{Data availability}
Restrictions apply to the availability of these data, which were used under license for the current study, and so are not publicly available.

\section*{Declarations}

\bmhead{Competing Interests}
The authors declare that they have no conflict of interest.

\bmhead{Ethical and informed consent for data used}
Not applicable.

\bibliography{references}

\begin{appendices}
\setcounter{table}{0}   
\setcounter{figure}{0}

\section{Details of evaluation datasets}\label{Appendix A}

This section details the datasets involved in the chinese telecom customer service evaluation benchmark.

\textit{\textbf{COTAS}} A homogeneous ticket recommendation task for change operation tickets in intelligent ticket creation scenarios. Given the title information of a new ticket, recommend similar historical change operation tickets. The test set size is 500, and results are evaluated using the F1 score.

\textit{\textbf{HNFTR}} A homogeneous ticket recommendation task for network fault tickets in intelligent ticket creation scenarios. Analyzes prompt information from new tickets to identify and recommend similar historical network fault tickets. The test set size is 151, and results are evaluated using the F1 score.

\textit{\textbf{OSEQA}} A knowledge Q\&A task for intelligent assistants in intelligent handling scenarios. Generates accurate and relevant answers based on staff queries and reference materials. The test set size is 87, and results are evaluated using Rouge-L.

\textit{\textbf{MIR-24}} A text classification task for maintenance personnel intent recognition in intelligent handling scenarios. Classifies input queries into one of 24 predefined categories. The test set size is 1653, and results are evaluated using accuracy.

\textit{\textbf{MIR-45}} A text classification task for maintenance personnel intent recognition in Intelligent Handling scenarios. Classifies input queries into one of 45 categories. The test set size is 1637, and results are evaluated using accuracy.

\textit{\textbf{MIR-83}} A fine-grained text classification task for maintenance personnel intent recognition with 83 categories. The test set size is 2000, and results are evaluated using accuracy.

\textit{\textbf{NFT-CRAG}} A ticket closure report generation task for network fault tickets in intelligent ticket closure scenarios. Given each network fault ticket information, analyze the fault phenomenon, handling process, and handling timeliness, and summarize them into a report. The test set size is 100, and results are evaluated using Rouge-L.

\textit{\textbf{HDRT-CRAG}} A ticket closure report generation task for hidden danger rectification tickets in intelligent ticket closure scenarios. Given each hidden danger rectification ticket information, analyze the overview of hidden dangers, rectification status, and handling timeliness, and summarize them into a report. The test set size is 100, and results are evaluated using Rouge-L.

\textit{\textbf{COT-CRAG}} A ticket closure report generation task for change operation tickets in intelligent ticket closure scenarios. Given each change operation ticket information, analyze the change information, impact assessment, and handling timeliness, and summarize them into a report. The test set size is 100, and results are evaluated using Rouge-L.

\textit{\textbf{MPWSG}} A daily work report generation task for maintenance personnel in intelligent ticket closure scenarios. Given the ticket information handled by installation and maintenance personnel in a day, analyze the workload, work type, and work efficiency, and summarize them into a report. The test set size is 100, and results are evaluated using Rouge-L.

\textit{\textbf{MIT-RA}} A structured operational report analysis task for maintenance issue tickets in intelligent evaluation scenarios. Given the statistical data of installation and maintenance problem tickets in one day, generate a structured operational report. The generated results are evaluated using a judger model.

\textit{\textbf{ITST-RA}} A structured operational report analysis task for IT support tickets in intelligent evaluation scenarios. Given the statistical data of IT support tickets in a month, generate a structured operational report. The generated results are evaluated using a judger model.

\textit{\textbf{TSGTV-RA}} A structured operational report analysis task for technical support group ticket volumes in intelligent evaluation scenarios. Given the statistical data of technical support group ticket volumes in a month, generate a structured operational report. The generated results are evaluated using a judger model.

\textit{\textbf{HDRT-RA}} A structured operational report analysis task for hidden danger rectification tickets in intelligent evaluation scenarios. Given the statistical data of hidden danger rectification tickets from January to October 2024, generate a structured operational report. The generated results are evaluated using a judger model.

\textit{\textbf{COT-RA}} A structured operational report analysis task for change operation tickets in intelligent evaluation scenarios. Given the statistical data of change operation tickets in a month, generate a structured operational report. The generated results are evaluated using a judger model.

\section{Prompts of evaluation datasets}\label{Appendix B}

\begin{figure}[b]
    \centering
    \includegraphics[width=0.5\textwidth]{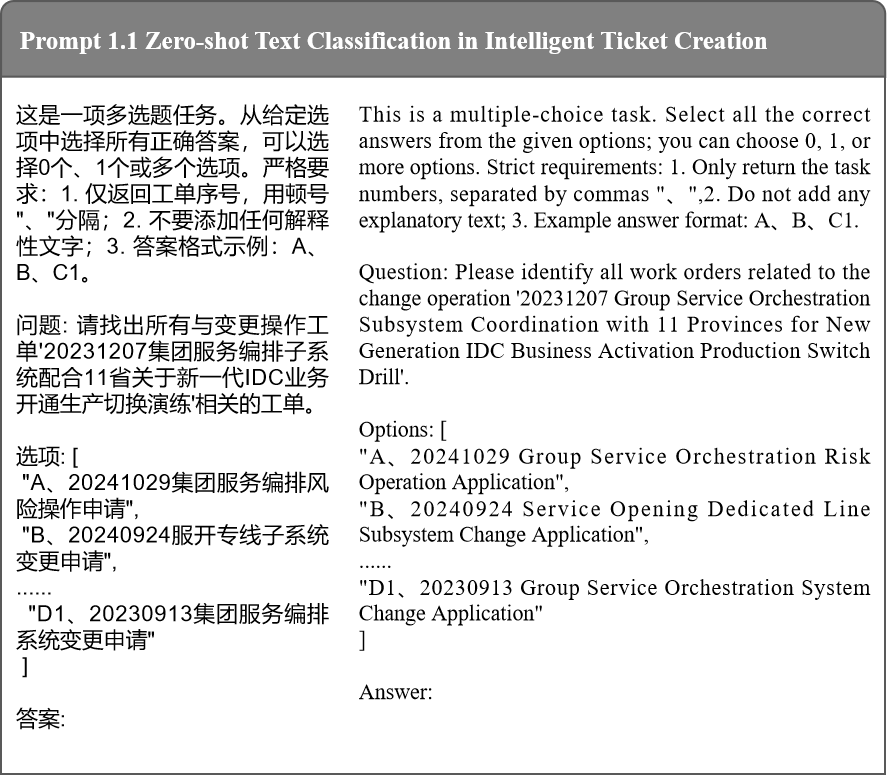}
    \hspace{2cm} 
    \includegraphics[width=0.5\textwidth]{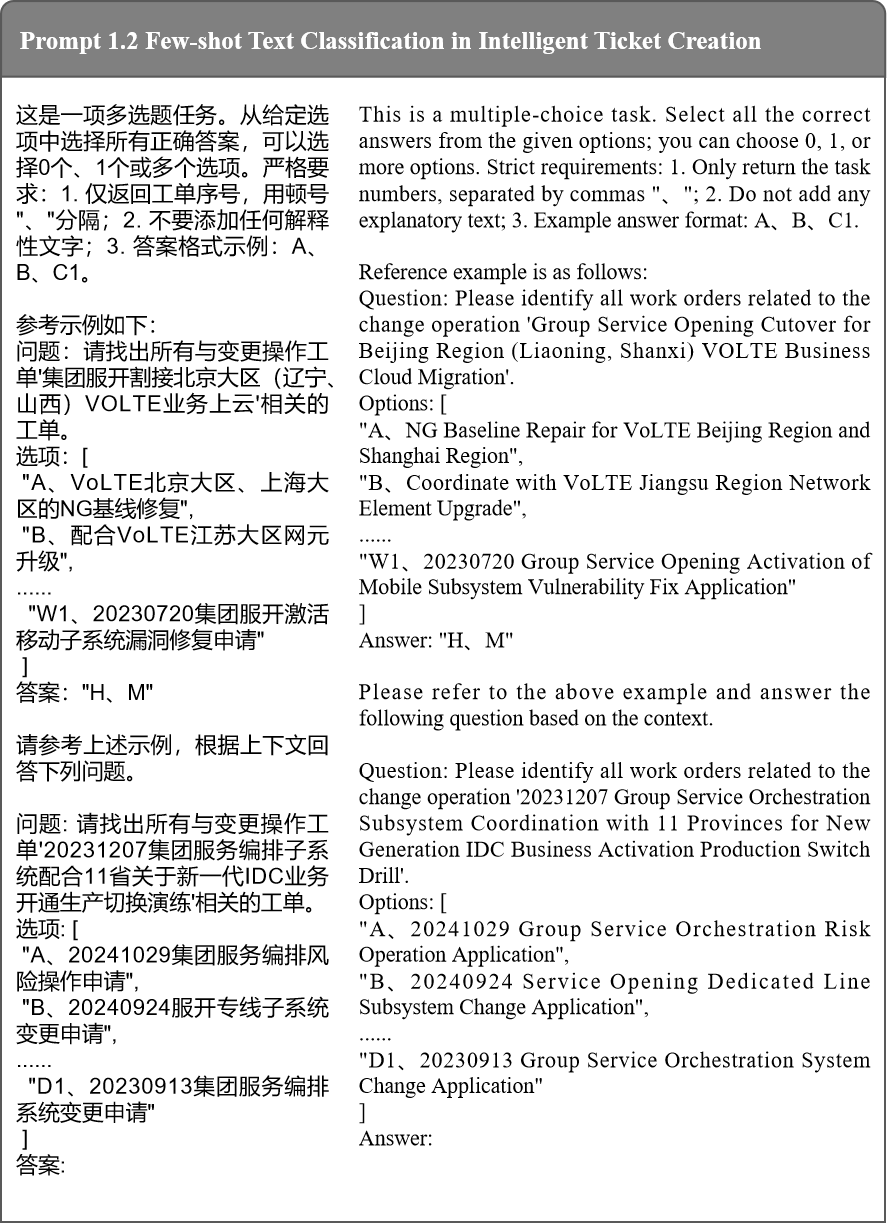}
    \caption{Zero-shot and few-shot prompt templates for constructing NLP task instructions in the intelligent ticket creation scenario (exemplified by the COTAS dataset), where the reference example is as follows: contains the samples.}
    \label{fig6}
\end{figure}

\begin{figure}[b]
    \centering
    \includegraphics[width=0.5\textwidth]{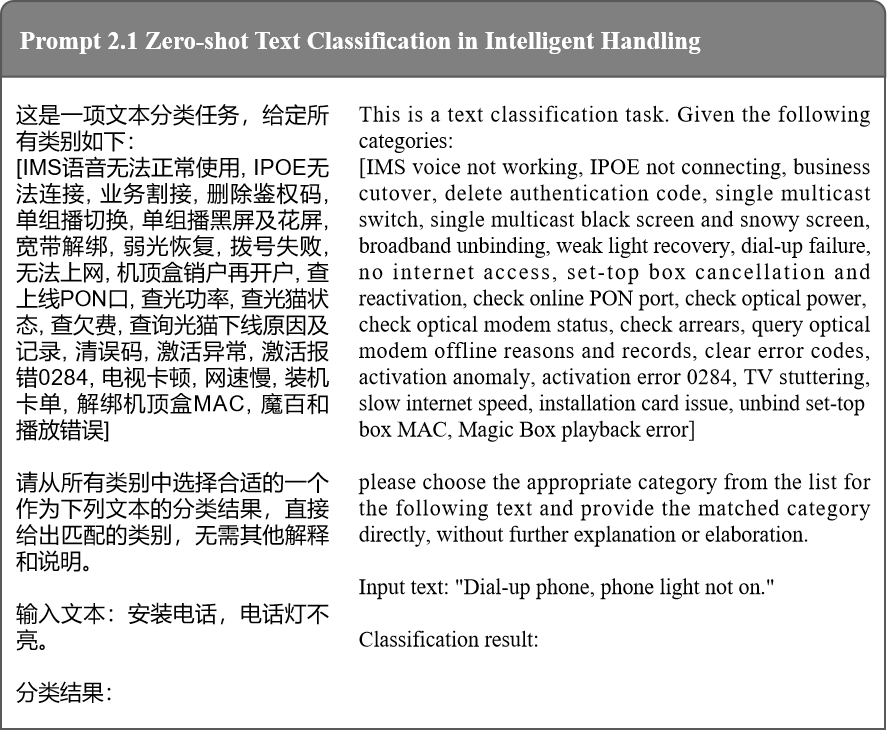}
    \hspace{2cm} 
    \includegraphics[width=0.5\textwidth]{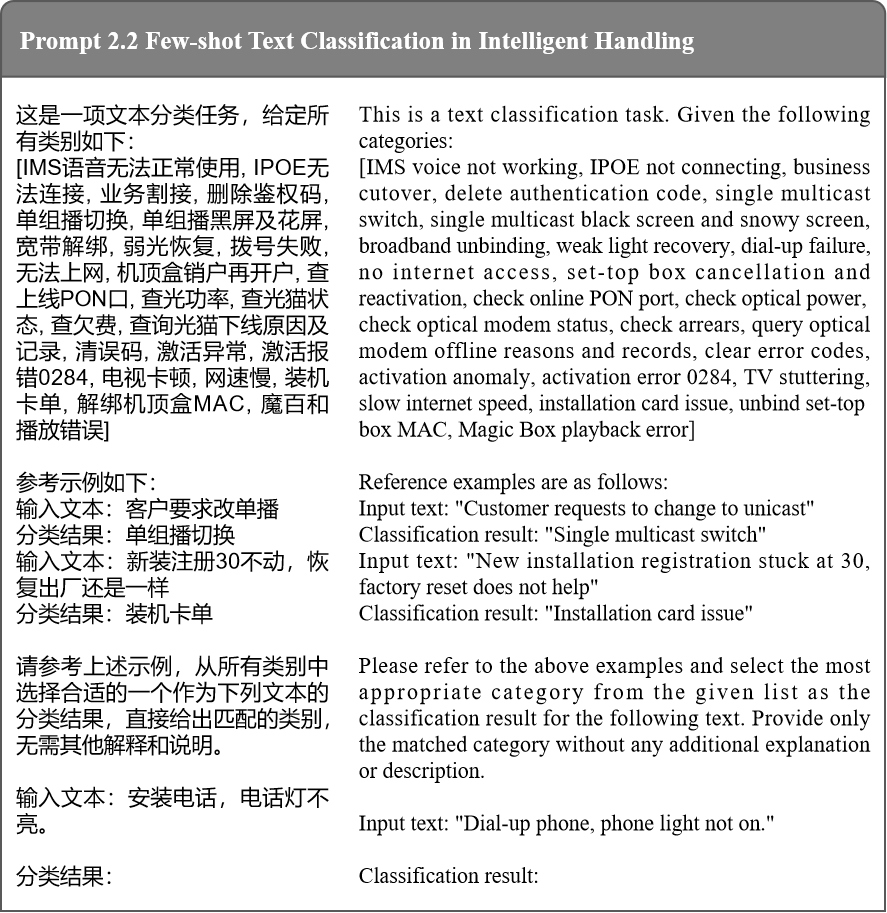}
    \caption{Zero-shot and few-shot prompt templates for constructing NLP task instructions in the intelligent handling scenario (exemplified by the MIR-24 dataset), where the reference example is as follows: contains the samples.}
    \label{fig7}
\end{figure}

\begin{figure}[b]
    \centering
    \includegraphics[width=0.5\textwidth]{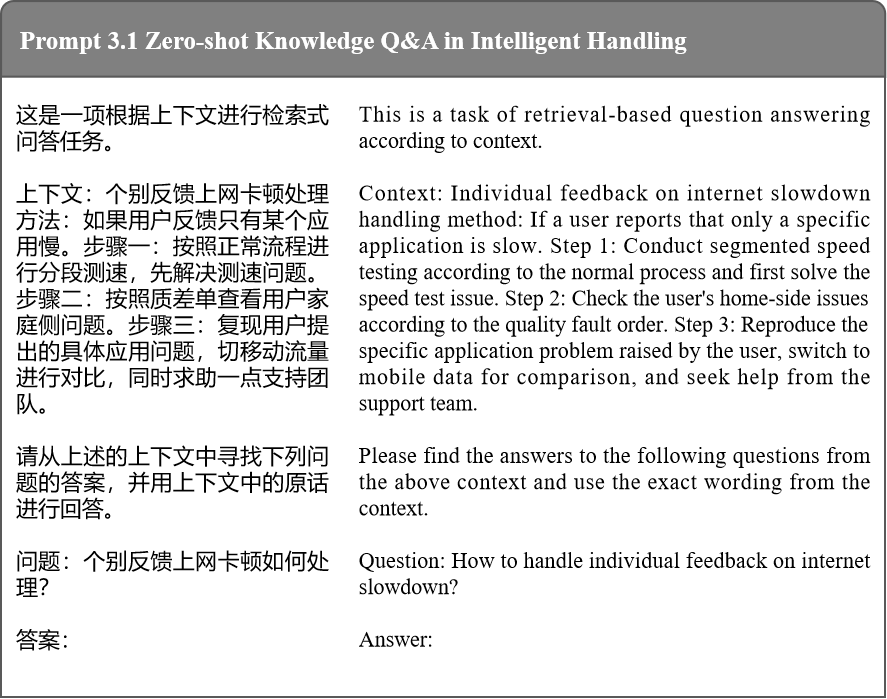}
    \hspace{2cm} 
    \includegraphics[width=0.5\textwidth]{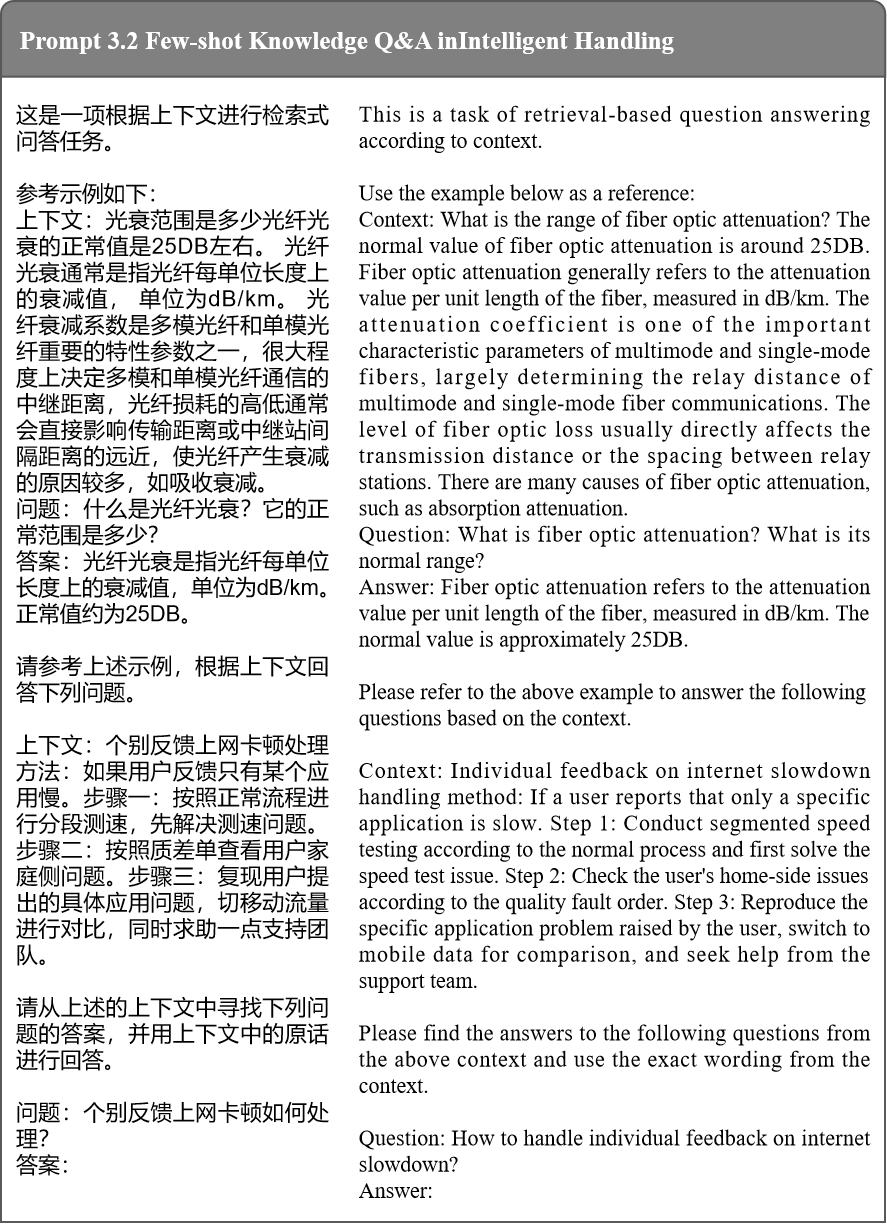}
    \caption{Zero-shot and few-shot prompt templates for constructing Knowledge Q\&A task instructions in the intelligent handling scenario (exemplified by the OSEQA dataset), where the reference example is as follows: contains the samples.}
    \label{fig8}
\end{figure}

\begin{figure}[b]
    \centering
    \includegraphics[width=0.5\textwidth]{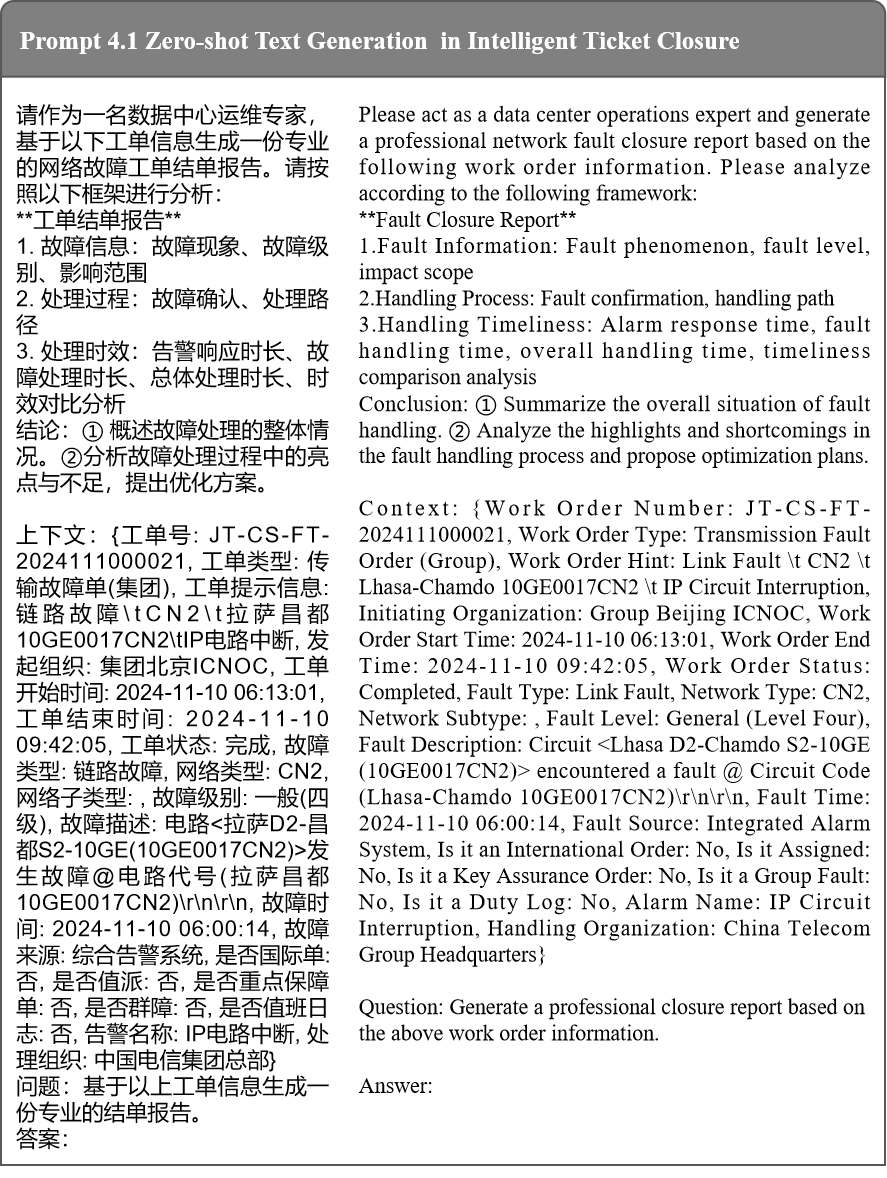}
    \hspace{2cm} 
    \includegraphics[width=0.5\textwidth]{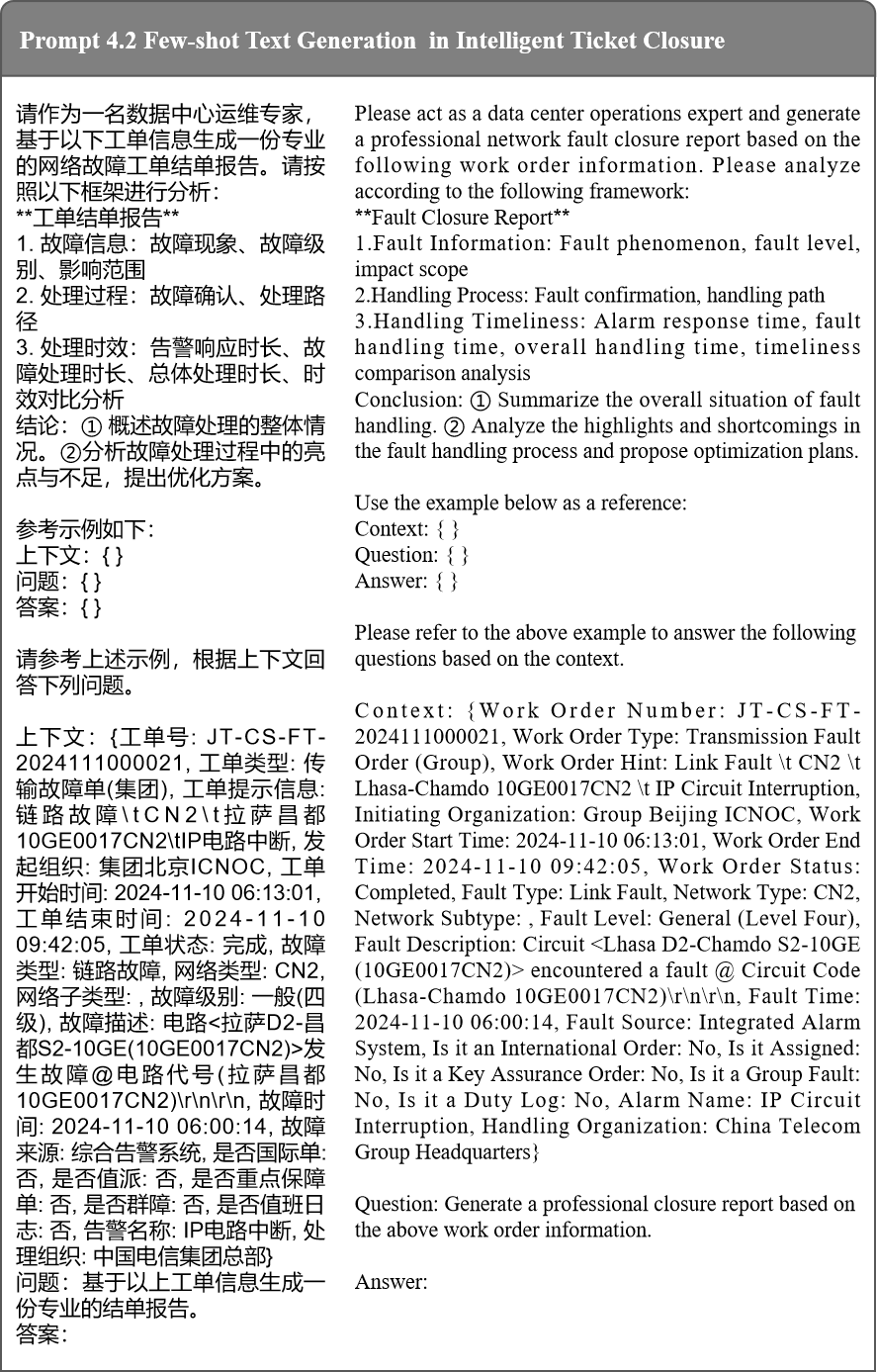}
    \caption{Zero-shot and few-shot prompt templates for constructing report generation task instructions in the intelligent ticket closure (exemplified by the NFT-CRAG dataset), where the reference example is as follows: contains the samples.}
    \label{fig9}
\end{figure}

\begin{figure}[b]
    \centering
    \includegraphics[width=0.5\textwidth]{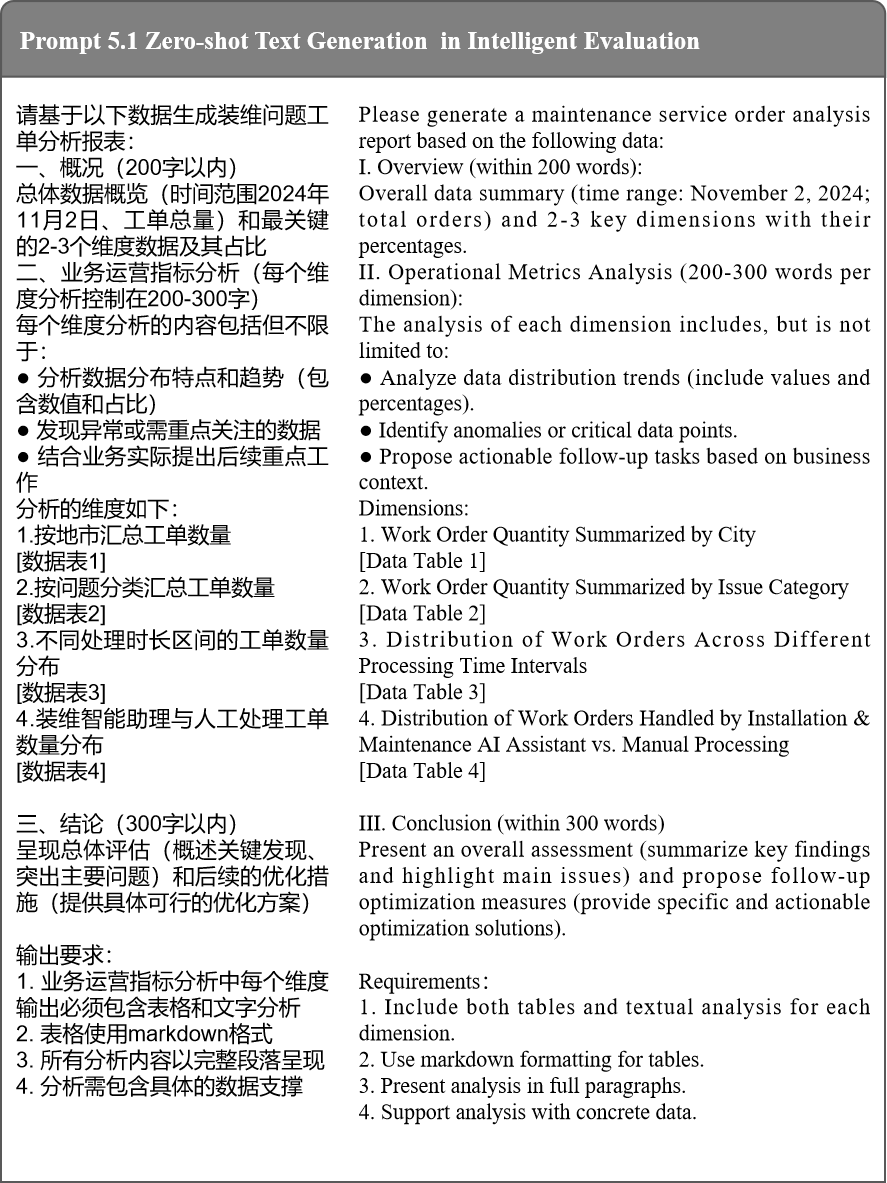}
    \caption{Zero-shot templates for constructing report analysis task instructions in the intelligent evaluation (exemplified by the MIT-RA dataset).}
    \label{fig10}
\end{figure}

The prompt templates used for text classification tasks in the intelligent ticket creation scenario are shown in Figure~\ref{fig6}. Prompt 1.1 is designed to generate answers when questions and options are provided, and Prompt 1.2 generates answers in a few-shot manner based on examples. This task includes the COTAS and HNFTR datasets, as illustrated in Table~\ref{tab1}. The prompt templates for text classification tasks in the intelligent handling scenario are shown in Figure~\ref{fig7}. Prompt 2.1 is used to generate answers when questions and classifications are provided, and Prompt 2.2 generates answers in a few-shot manner based on examples. This task involves the MIR-24, MIR-45, and MIR-83 datasets. The zero-shot and few-shot prompts for the knowledge question-answering task in the intelligent handling scenario are presented in Figure~\ref{fig8}. The dataset for this task is OSEQA. The zero-shot and few-shot prompts for the text generation task in the intelligent ticket closure scenario are shown in Figure~\ref{fig9}. This task includes four datasets: NFT-CRAG, HDRT-CRAG, COT-CRAG, and MPWSG. The zero-shot prompt for the knowledge question-answering task is demonstrated in Prompt 5.1 of Figure~\ref{fig10}, which provides a report framework. This task includes five datasets: MIT-RA, ITST-RA, TSGTV-RA, HDRT-RA, and COT-RA.

\twocolumn

\end{appendices}

\end{document}